\documentclass[]{article}

\usepackage[final]{neurips_2022}

\usepackage{soul}
\usepackage{url}
\usepackage{epstopdf}
\usepackage{float}
\usepackage[utf8]{inputenc}
\usepackage{graphicx}
\usepackage{amsmath}
\usepackage{amssymb}
\usepackage{booktabs}
\usepackage{algorithm}
\usepackage[noend] {algorithmic}
\usepackage{enumitem}
\usepackage{booktabs} 
\usepackage{graphicx,array}
\usepackage{color,soul,xspace}

\usepackage{comment}
\usepackage{epsfig}

\usepackage{caption}
\usepackage{subcaption}

\usepackage{mathrsfs,mdwlist,enumitem}

\newcommand{\gnns}{\textsc{Gnn}s\xspace}
\newcommand{\gnn}{\textsc{Gnn}\xspace}
\newcommand{\fgn}{\textsc{FGnn}\xspace}

\newcommand{\lnn}{\textsc{Lnn}\xspace}
\newcommand{\hnn}{\textsc{Hnn}\xspace}

\newcommand{\MLP}{\texttt{MLP}\xspace}
\newcommand{\sqp}{\texttt{squareplus}\xspace}

\newcommand{\CU}{\mathcal{U}\xspace}

\newcommand{\CV}{\mathcal{V}\xspace}
\newcommand{\CE}{\mathcal{E}\xspace}

\newcommand{\cW}{\mathbf{W}\xspace}

\newcommand{\cx}{\mathbf{x}\xspace}
\newcommand{\cy}{\mathbf{y}\xspace}
\newcommand{\ch}{\mathbf{h}\xspace}

\newcommand{\cz}{\mathbf{z}\xspace}

\newcommand{\m}{\textbf{M}\xspace}
\newcommand{\C}{\textbf{C}\xspace}
\newcommand{\f}{\textbf{F}\xspace}

\newcommand{\x}{\textbf{x}\xspace}
\newcommand{\p}{\textbf{p}_\textbf{x}\xspace}
\newcommand{\pdot}{\dot{\textbf{p}}_\textbf{x}\xspace}
\newcommand{\xdot}{\dot{\textbf{x}}\xspace}
\newcommand{\xddot}{\ddot{\textbf{x}}\xspace}
\newcommand{\node}{\textsc{Node}\xspace}
\newcommand{\gnode}{\textsc{Gnode}\xspace}
\newcommand{\lgn}{\textsc{Lgn}\xspace}
\newcommand{\hgn}{\textsc{Hgn}\xspace}
\newcommand{\fgnn}{\textsc{Fgnn}\xspace}
\newcommand{\lgnn}{\textsc{Lgnn}\xspace}
\newcommand{\hgnn}{\textsc{Hgnn}\xspace}
\newcommand{\fgnode}{\textsc{FGnode}\xspace}
\newcommand{\clgn}{\textsc{CLgn}\xspace}
\newcommand{\chgn}{\textsc{CHgn}\xspace}
\newcommand{\clgnn}{\textsc{CLgnn}\xspace}
\newcommand{\chgnn}{\textsc{CHgnn}\xspace}
\newcommand{\cfgnode}{\textsc{CFGnode}\xspace}

\newcommand{\cgnode}{\textsc{CGnode}\xspace}

\setlist{nolistsep,leftmargin=*}

\newcommand{\rev}[1]{\textcolor{black}{#1}}

\usepackage[utf8]{inputenc} 
\usepackage[T1]{fontenc}    
\usepackage{hyperref}       
\usepackage{url}            
\usepackage{booktabs}       
\usepackage{amsfonts}       
\usepackage{nicefrac}       
\usepackage{microtype}      
\usepackage{xcolor}         
\usepackage{graphicx}
\usepackage{wrapfig}

\title{Unravelling the Performance of Physics-informed Graph Neural Networks for Dynamical Systems}

\author{%
  Abishek Thangamuthu, Gunjan Kumar\\
  Department of Computer Science and Engineering\\
  Indian Institute of Technology Delhi\\
  New Delhi 110016, India\\
  \And
  Suresh Bishnoi \\
  School of Interdisciplinary Research\\
  Indian Institute of Technology Delhi\\
  New Delhi 110016, India\\
  \And
  Ravinder Bhattoo \\
  Department of Civil Engineering\\
  Indian Institute of Technology Delhi\\
  New Delhi 110016, India\\
  \And
  N M Anoop Krishnan, Sayan Ranu \\
  Yardi School of Artificial Intelligence \\
  Indian Institute of Technology Delhi\\
  New Delhi 110016, India\\
  \texttt{\{krishnan,sayanranu\}@iitd.ac.in}
}

\begin{document}

\maketitle

\begin{abstract}
\vspace{-0.10in}
Recently, graph neural networks have been gaining a lot of attention to simulate dynamical systems due to their inductive nature leading to \textit{zero-shot generalizability}. Similarly, physics-informed inductive biases in deep-learning frameworks have been shown to give superior performance in learning the dynamics of physical systems. There is a growing volume of literature that attempts to combine these two approaches. Here, we evaluate the performance of \rev{thirteen} different graph neural networks, namely, \textit{Hamiltonian} and \textit{Lagrangian} graph neural networks, graph \textit{neural ODE}, and their variants with explicit constraints and different architectures. We briefly explain the theoretical formulation highlighting the similarities and differences in the inductive biases and graph architecture of these systems. We evaluate these models on spring, pendulum, \rev{gravitational, and 3D deformable solid} systems to compare the performance in terms of rollout error, conserved quantities such as energy and momentum, and generalizability to unseen system sizes. Our study demonstrates that \gnn{s} with additional inductive biases, such as explicit constraints and decoupling of kinetic and potential energies, exhibit significantly enhanced performance. Further, all the physics-informed \gnn{s} exhibit zero-shot generalizability to system sizes an order of magnitude larger than the training system, thus providing a promising route to simulate large-scale realistic systems.
\looseness=-1
\end{abstract}

\vspace{-0.20in}
\section{Introduction and Related Works}
\label{sec:intro}
\vspace{-0.10in}
Understanding the time evolution or ``dynamics'' of physical systems is a long-standing problem of interest with both fundamental and practical relevance in areas of physics, engineering, mathematics, and biology~\cite{goldstein2011classical,karniadakis2021physics,lavalle2006planning,murray2017mathematical,park2021accurate,noether1971invariant}. Traditionally, the dynamics of physical systems are expressed in terms of differential equations with respect to their state variables, such as the position ($\textbf{x}(t)$) or velocity ($\dot{\cx}(t)$)~\cite{goldstein2011classical,lavalle2006planning}. Note that the state variables are actual observable quantities and define the configurational state of a system at any point of time $t$. The differential equation is then solved with the initial and boundary conditions to predict the future states of the system. While this conventional approach is highly efficient and requires little data to predict the dynamics, in many cases the exact differential equation may be unknown. Further, the formulation of the differential equations might require the knowledge of abstract quantities such as the energy, or force distribution of the system, which are not directly measurable in most cases~\cite{cranmer2020discovering}.

To this extent, neural networks or MLPs present as efficient function approximators, that can learn the dynamics directly from the state or trajectory~\cite{karniadakis2021physics,chen2018neural}. The learned dynamics can then be used to infer the future states of the system with different initial conditions. It has been shown recently that the learning can be significantly enhanced if \textit{physics-based inductive biases} are provided to these MLPs~\cite{lnn,lnn1,roehrl2020modeling,bhattoo2022learning,bishnoi2022enhancing}. Specifically, these biases allow the MLP to preserve the characteristics of physical systems, such as energy and momentum conservation, and thus lead to a realistic realization of a trajectory of the system~\cite{greydanus2019hamiltonian,lnn,zhong2020unsupervised}. The most popular choices for learning dynamics are the \textit{neural ODE} (\node)~\cite{chen2018neural,gruver2021deconstructing,bishnoi2022enhancing},\textit{Lagrangian} (\lnn{s})~\cite{lnn,lnn1,lnn2,bhattoo2022learning}, and \textit{Hamiltonian} neural networks \rev{(\hnn{s})}~\cite{sanchez2019hamiltonian,greydanus2019hamiltonian,zhong2020dissipative,zhong2019symplectic}. While \node learns the differential equations as a parameterized neural network~\cite{chen2018neural}, \lnn and \hnn learn the \textit{Lagrangian} and \textit{Hamiltonian} of the system, respectively, which is then used to predict the forward trajectory through the physics-based equation. It is worth noting that in all these cases, the training is purely based on the trajectory or state of the system~\cite{haitsiukevich2022learning}. 

One of the major disadvantages of these systems is that they are \textit{transductive} in nature, that is, they work only for the systems they are trained for. For instance, an \lnn trained for a double pendulum can be used to infer the dynamics of a double pendulum only and not any $n$-pendulum. This significantly limits the application of \lnn, \hnn, and \node to simple systems since for each system the training trajectory needs to be generated and the model needs to be trained. It has been shown the transductivity of MLPs could be addressed by incorporating an additional inductive bias in the structure by accounting for the topology of the system in the graph framework using a graph neural network (\textsc{Gnn})~\cite{battaglia2016interaction,sanchez2020learning,cranmer2020discovering,sanchez2019hamiltonian,bhattoo2022learning,bishnoi2022enhancing}. \textsc{Gnn}s, once trained has the capability to generalize to arbitrary system sizes. 
\looseness=-1

\begin{figure}[t]
    \centering
    \begin{subfigure}{0.49\textwidth}
        \centering
        \includegraphics[width=0.99\textwidth]{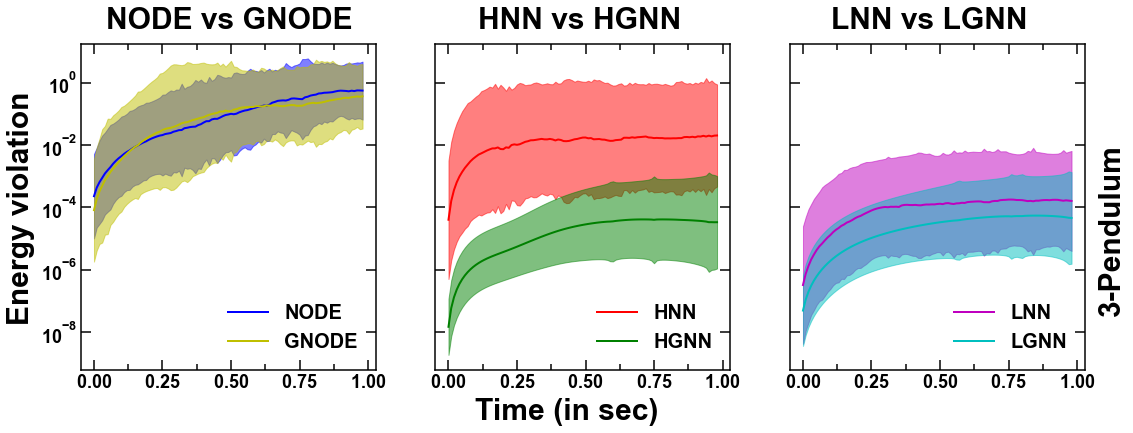}
    \end{subfigure}\hfill
    \begin{subfigure}{0.49\textwidth}
        \centering
        \includegraphics[width=0.99\textwidth]{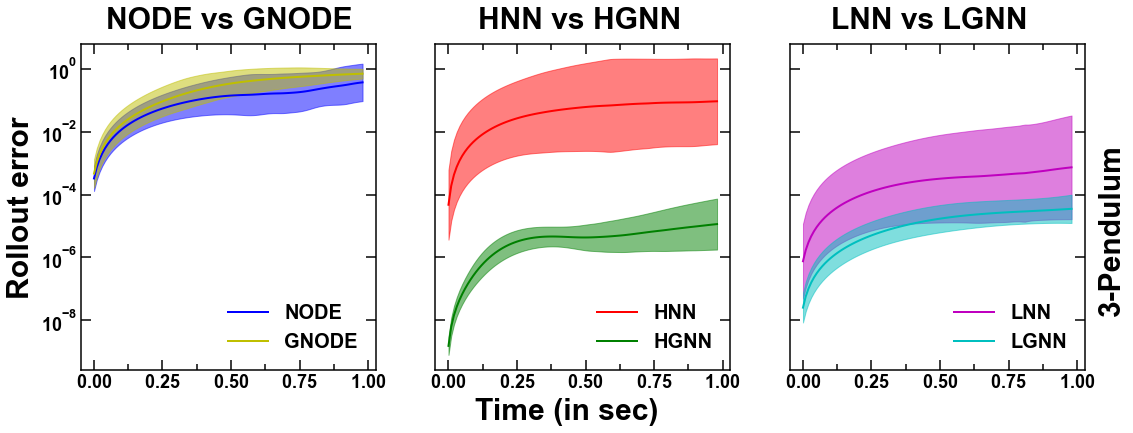}
    \end{subfigure}
    \vspace{-0.05in}
    \caption{Comparison of physics-informed neural networks (\node, \hnn, and \lnn) and their graph-counterparts (\gnode, \hgnn, \lgnn) for 3-pendulum systems. 
    The error bar represents the $95\%$ confidence interval over $100$ trajectories generated from random initial conditions. Further details of the architectures, and the datasets are discussed in \S~\ref{sec:models} and \S~\ref{sec:datasets} respectively.}
    \label{fig:graphvsmlp}
    \vspace{-0.25in}
\end{figure}

\rev{It is worth noting that most of the studies on \gnn{s} for dynamical systems use a purely data-driven approach, where the \gnns are used to learn the position and velocity update directly from the data.} Recently, physics-informed \textsc{Gnn}s have been used to simulate complex physical systems~\cite{sanchez2020learning}. In addition, since the \gnn{s} are trained at the node and edge level, \rev{they} can potentially learn more efficiently from the same number of data points in comparison to their fully connected counterparts. Figure~\ref{fig:graphvsmlp} shows the energy and rollout error of \node, \lnn, and \hnn with their graph-based counterparts, namely, \gnode, \lgnn, and \hgnn (detailed structure discussed in Section~\ref{sec:models}) for 3-pendulum systems. We observe that the graph-based versions provide comparable or better performance than their counterparts when trained on the same amount of data for the same number of epochs. While this difference is not significant in \node, in \lgn and \hgn, we observe that the difference is more pronounced. In addition, as demonstrated later in this work (see Section~\ref{sec:generalizability}), the graph-versions have the additional advantage of inductivity to generalize to larger system sizes.


\rev{
Despite the wide use of \gnn{s} to model physical systems, there exist no work thus far that systematically benchmark their performance for learning and inferring the dynamics of physical systems. In this work, we aim to address this gap by evaluating the role of different physics-based inductive biases in \gnn{s} for learning the dynamics of complex systems such as $n$-spring and pendulum systems, gravitational bodies, and elastically deformable 3D solid}. The major contributions of the present work are as follows. 
\vspace{-0.05in}
\begin{enumerate}
\item {\bf Topology-aware modeling.} We benchmark \rev{thirteen} different physics-informed \gnn{s} that model physical systems. By analyzing their energy, rollout, and momentum violation, we carefully evaluate and benchmark their performance. 
\rev{\item{\bf Physics-informed inductive bias.} We analyze the nature of inductive bias provided for \gnns by Lagrangian, Hamiltonian and Newtonian mechanics in learning the dynamics, although they, in principle, are equivalent formulations.}
\item{\bf Explicit constraints.} We analyze the role of providing explicit constraints as an inductive bias on the performance of different \gnn{s}.
\item{\bf Decoupling potential and kinetic energies.} We analyze the effect of exploiting the Hamiltonian and Lagrangian structure, which allows decoupling of kinetic and potential energies in the Cartesian coordinates. Specifically, we evaluate how this decoupling, by modifying the graph architecture, affect the performance of \gnn{s}.
\item{\bf Zero-shot generalizability.} Finally, we evaluate the ability of these \gnn{s} to generalize to arbitrary system sizes an order of magnitude larger than the systems they have been trained on. 
\end{enumerate}
\vspace{-0.1in}
\section{Preliminaries on Learning Dynamical Systems}
\label{sec:prelim}
\vspace{-0.1in}
We consider a rigid body system comprising of $n$ interacting particles. The configuration of this system is represented as a set of Cartesian coordinates $\x(t)=(x_1(t), x_2(t),\ldots,x_n(t))$.
Since we are using a graph neural network to model the physical interactions, it is natural to select the Cartesian coordinates for the features such as position as velocity. While this may result in an increased complexity in the form of the Hamiltonian or Lagrangian, it significantly simplifies the mass matrix for particle-based systems by making it positive definite~\cite{lnn1} and diagonal~\cite{lnn1,zhong2021benchmarking}.
\vspace{-0.1in}
\subsection{ODE Formulation of Dynamics}
\label{sec:ode}
\vspace{-0.1in}
Traditionally, the dynamics of a system can be expressed in terms of the \textit{D'Alembert's principle} as~\cite{lavalle2006planning}
\begin{equation}
\m \xddot - \f(\x,\xdot) = 0
\label{eq:ode}
\end{equation}
where, in Cartesian coordinates, $\m$ is the constant mass matrix that is independent of the coordinates~\cite{zhong2021benchmarking}, and $\f$ represents the dynamics of the system. Accordingly, the acceleration $\xddot$ of the system can be computed as:
\vspace{-0.15in}
\begin{equation}
\label{eq:acc_node}
    \xddot = \m^{-1} (\f(\x,\xdot))
\end{equation}

In case of systems with constraints, for instance a pendulum system where the length between the two bobs are maintained constant, Eq.~\ref{eq:ode} can be modified to feature the constraints explicitly~\cite{lavalle2006planning,murray2017mathematical}.
\vspace{-0.05in}
\begin{equation}
    \m \xddot - \f(\x) + A^T\lambda = 0
\label{eq:node_decoupled}
\end{equation} 
Here, the constraints of the system are given by $A(\x)\xdot=0$, where $A(\x) \in \mathbb{R}^{k\times D}$ represents $k$ constraints \rev{associated with the $D$ degrees of freedom (See App.~\ref{app:constraints} for details on why constraints are expressed through this form)}. Thus, $A$ represents the non-normalized basis of constraint forces given by $A=\nabla_\x (\phi)$, where $\phi$ is the matrix corresponding to the constraints of the system and $\lambda$ is the \textit{Lagrange multiplier} corresponding to the relative magnitude of the constraint forces. 

To solve for $\lambda$, the constraint equation can be differentiated with respect to time to obtain $A\xddot+\dot{A}\xdot=0$. Substituting for $\xddot$ from Eq.~\ref{eq:node_decoupled} and solving for $\lambda$, we get: 
\begin{equation}
    \lambda = (A \m^{-1}A^T)^{-1}(A \m^{-1}(\f ) + \dot A \xdot)
\end{equation}
Accordingly, $\xddot$ can be obtained as:
\begin{equation}
    \xddot = \m^{-1}\left( \f -  A^T (A 
    \m^{-1}A^T)^{-1}\left(A \m^{-1}( \f) + \dot A \xdot \right)\right)
    \label{eq:acc_code}
\end{equation}

\vspace{-0.2in}
\subsection{Lagrangian Dynamics}
\vspace{-0.1in}
The standard form of Lagrange's equation for a system with $holonomic$ constraints is given by
\begin{equation}
    \frac{d}{dt} \left( \nabla_{\xdot} L\right)-\left( \nabla_{\x} L \right) = 0
\end{equation}
where the Lagrangian is $L(\x,\xdot,t)=T(\x,\xdot,t)-V(\x,t)$ with $T(\x,\xdot,t)$ and $V(\x,t)$ representing the total kinetic energy of the system and the potential function from which generalized forces can be derived, respectively. Accordingly, the dynamics of the system can be represented using \textit{Euler-Lagrange (EL)} equations as
\vspace{-0.05in}
\begin{equation}
\xddot = \left( \nabla_{\xdot \xdot} L \right)^{-1} \left[ \nabla_{\x} L - \left( \nabla_{\xdot{\x}} L \right) \xdot\right]
\label{eq:EL}
\end{equation}

\rev{Here, $\nabla_{\dot{\cx}\dot{\cx}}$ refers to $\frac{\partial^2}{\partial \dot{\cx}^2}$}. In systems with constraints, the Lagrangian formulation can be modified to include the explicit constraints~\cite{lavalle2006planning}. Accordingly, the acceleration can be computed as
\begin{equation}
    \xddot = \nabla_{\xdot \xdot} L \left(- \nabla_{\xdot{\x}} L \xdot - \nabla_{\x} L -  A^T (A 
    (\nabla_{\xdot \xdot} L)^{-1}A^T)^{-1}\left(A (\nabla_{\xdot \xdot} L)^{-1}(- \nabla_{\xdot{\x}} L \xdot - \nabla_{\x} L) + \dot A \xdot \right)\right)
    \label{eq:acc_lgn}
\end{equation}

In Cartesian coordinates, the Lagrangian simplifies to $L(\x,\xdot)=\frac{1}{2}\xdot^T\m\xdot-V(\x)$. Exploiting the structure of Lagrangian by decoupling the kinetic and potential energies, and substituting this expression in Eq.~\ref{eq:acc_lgn}, we obtain $\m = \nabla_{\xdot \xdot} L$ as a constant matrix independent of coordinates, $\C = \nabla_{\xdot{\x}} L = 0$, and $\f = \nabla_{\x} V(\x)$. Accordingly, the $\xddot$ can be obtained as
\begin{equation}
    \xddot = \m^{-1}\left(- \nabla_{\x} V(\x) -  A^T (A 
    \m^{-1}A^T)^{-1}\left(A \m^{-1}(-\nabla_{\x} V(\x)) + \dot A \xdot \right)\right)
    \label{eq:acc_clgnn}
\end{equation}

\vspace{-0.1in}
\subsection{Hamiltonian Dynamics} 
\label{sec:HD}
\vspace{-0.1in}
Hamiltonian equations of motion are given by
\begin{equation}
    {\x = \nabla_{\p} H}, \qquad {\pdot = -\nabla_\x H}
\end{equation}
where, $\p = \nabla_{\xdot} L = \m \xdot$ represents the momentum of the system in Cartesian coordinates and $H(\x,\p) = \xdot^T\p-L(\x,\xdot) = T(\xdot) + V(\x)$ represents the Hamiltonian of the system. The equation can be simplified by assuming $Z = [\x; \p]$ and $J = [0, I ; -I , 0]$ then the Hamiltonian equation can be written as
\begin{equation}
    \nabla_Z H + J\dot{Z} = 0
    \label{eq:ham}
\end{equation}
Note that this coupled first order differential equations are equivalent to the Lagrangian Eq.~\ref{eq:EL}.

In systems with constraints, the Hamiltonian equations of motion can be modified to feature the constraints explicitly, similar to the Lagrangian and ODE formulations. Hamiltonian equations with explicit constraints can be written as~\cite{zhong2021benchmarking}
\begin{equation}
\label{eq:ham_constraint}
    \nabla_Z H + J\dot{Z} + (D_Z \Psi)^T\lambda = 0
\end{equation}
where, $(D_Z \Psi)^T\lambda$ represents the effect of constraints on both the equations. Here, $\Psi(Z) = (\Phi;\dot{\Phi})$ and $\Phi = \Phi(\x)=0$ represent the constraints equation, which implies $\Psi(Z) = 0$. Thus, $(D_Z\Psi)\dot{Z}=0$. Substituting for $\dot{Z}$ and solving for $\lambda$ yields
\begin{equation}
\label{eq:lambda}
    \lambda = -[(D_Z\Psi)J(D_Z\Psi)^T]^{-1}[(D_Z\Psi)J(\nabla H)]
\end{equation}

Substituting this value into the Eq.~\ref{eq:ham_constraint} and solving for $\dot{Z}$ yields
\begin{equation}
    {\dot{Z}} = J[\nabla_Z H -(D_Z \Psi)^T [(D_Z\Psi)J(D_Z\Psi)^T]^{-1} (D_Z \Psi) J \nabla_Z H]
    \label{eq:acc_ham}
\end{equation}
\vspace{-0.2in}
\rev{\subsection{Physics-informed \gnns}}
\label{sec:pgnns}
\vspace{-0.10in}
\rev{Data-driven \gnns take a given configuration as input and predict the updated configuration~\cite{battaglia2016interaction}. Instead, physics-informed \gnns take the configuration (position and velocity) as input and predict abstract quantities such as \textit{Lagrangian} or \textit{Hamiltonian} or force. These output values are then used along with physics-based equations (as defined in \S~\ref{sec:ode}-\ref{sec:HD}) to obtain the updated trajectory. Thus, in physics-informed \gnns, the neural network essentially learns the function relating the position and velocity to quantities such as force or energy, by training on trajectory. }
\section{Models Studied}
\label{sec:models}
\vspace{-0.1in}
\begin{figure}[t]
\vspace{-0.30in}
    \centering
    \includegraphics[width=5in]{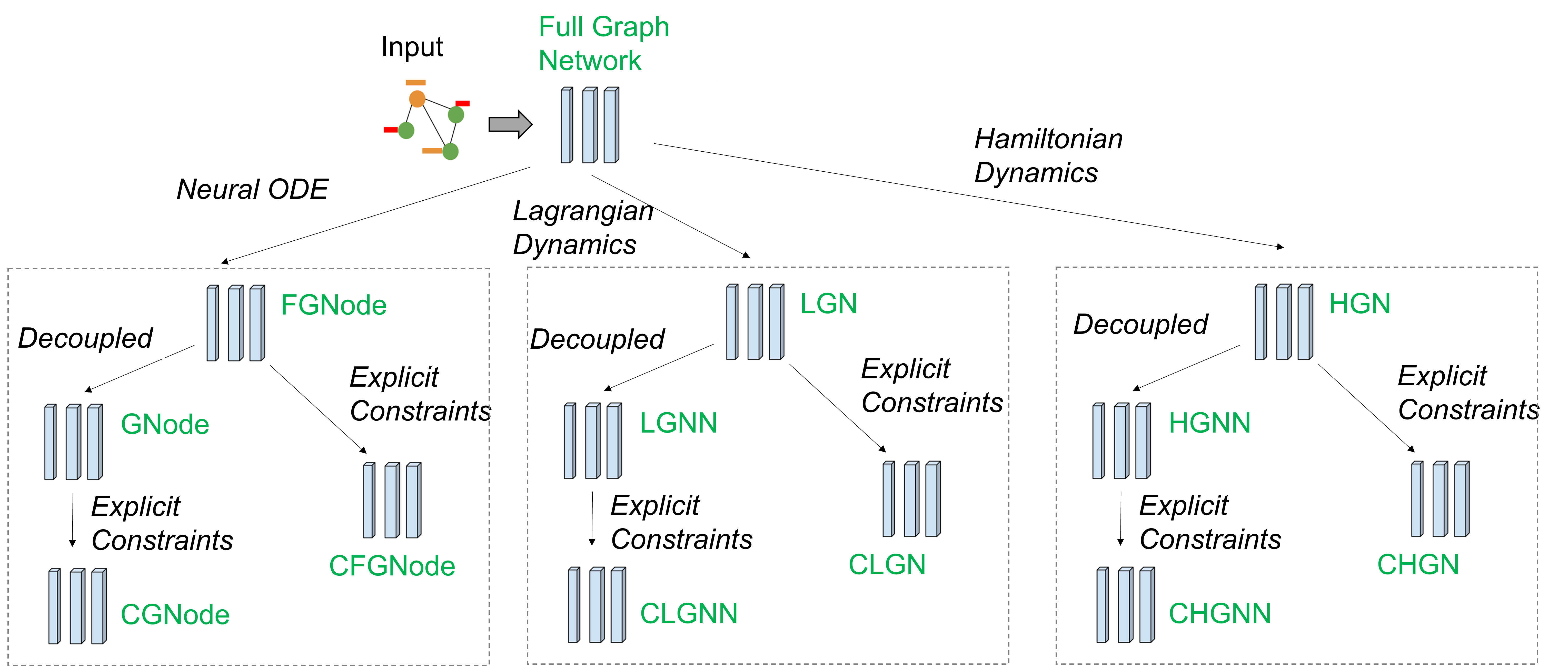}
    \caption{An overview of the various models being benchmarked and the relationship between them. Here, ``Decoupled'' indicates the model variety where the  kinetic and potential energies are decoupled with the force being a function of only the 
position and the mass matrix being diagonal in nature.}
    \label{fig:modelheirarchy}
     \vspace{-0.20in}
\end{figure}

In this section, we describe the models studied in this work. An overview of the models and the relationship between them is presented in Fig.~\ref{fig:modelheirarchy}. \rev{Specifically, the full graph network~\cite{sanchez2020learning} lies at the root, i.e., the base \gnn model}. On this basic framework, we study the impact of physics-based inductive biases in the form of Neural ODE, Lagrangian and Hamiltonian dynamics. Each of these inductive biases is further refined through the injection of explicit constraints and ``decoupled'' modeling of the potential and kinetic energies. We next explain each of these specific models. Note that the detailed architecture of the models are not discussed since they are available in the original papers that introduced the works. Accordingly, we mention the names of the models used in this study, and their major features.

$\bullet$ \textbf{Full Graph Network (\fgn)}~\cite{sanchez2020learning}: \fgn is a \textit{message-passing} \gnn, wherein each node $v$ (and potentially edges as well) draws ``messages'' from its $1$-hop neighbors (i.e., their representations in the form of feature vectors), and passes these messages through a multi-layer perceptron (\MLP) to construct $v$'s own representation. An \fgn of $\ell$ layers, therefore, learns node representations that capture the topology of the $\ell$-hop neighborhood around each node. \fgn attempts to directly predict the property being modeled (such as acceleration) without exploiting any physics-informed inductive biases.
\looseness=-1

$\bullet$ \textbf{\fgnode}\cite{sanchez2019hamiltonian}: \fgnode refers to a neural ODE version of \fgn. In \fgnode, the particle positions and velocities are given as the input to each node, while the difference in the position of nodes is given as a feature to each edge. The node-level output of the model is the acceleration on each node, which is then integrated to obtain the updated position and velocity using Eq.~\ref{eq:acc_node}. 

$\bullet$ \textbf{\gnode:} \gnode is equivalent to \fgnode with a minor modification in the architecture~\cite{bishnoi2022enhancing}. Specifically, we note that in particle-based systems in Cartesian coordinates, the kinetic and potential energies can be  \textit{decoupled} with the force being a function of only the position and the mass matrix being diagonal in nature~\cite{lnn1,lnn2,zhong2021benchmarking,bishnoi2022enhancing}. Thus, for \gnode, only the position of the particle is given as a node input and not the velocity, which is then used to predict the node-level acceleration using Eq.~\ref{eq:acc_node}. Further details of this architecture is provided in App.~\ref{app:gnode}.
 
$\bullet$ \textbf{\cfgnode and \cgnode:} \cfgnode and \cgnode are \fgnode and \gnode with explicit constraints~\cite{lnn1} as given by Eq.~\ref{eq:acc_code}, respectively.


$\bullet$ \textbf{Lagrangian Graph Network (\lgn)~\cite{lnn}:} Here, the graph architecture, which is an \fgn, directly predicts Lagrangian of the system. The acceleration is then predicted from the Lagrangian using \textit{Euler-Lagrange} equation (Eq.~\ref{eq:acc_lgn}).

$\bullet$ \textbf{\lgnn:} \lgnn \rev{improves on} \lgn, where the kinetic and potential energies are decoupled~\cite{bhattoo2022learning}. Further, the potential energy is predicted using a \gnn with the position as the node input and distance between the nodes as the edge input. The diagonal nature of mass matrix in particle-based rigid body systems in Cartesian coordinates is exploited to learn the parametric masses~\cite{lnn,lnn1,lnn2}. Further details of this architecture is provided in App.~\ref{app:lgnn}. 

$\bullet$ \textbf{\clgn and \clgnn:} \clgn and \clgnn refers to \lnn and \lgnn with explicit constraints respectively. Specifically, the acceleration is computed using Eq.~\ref{eq:acc_clgnn}~\cite{bhattoo2022learning,lavalle2006planning,lnn1}.

$\bullet \textbf{\hgn:}$ \hgn refers to Hamiltonian graph network, where the Hamiltonian of a system is predicted using the full graph network~\cite{sanchez2020learning,sanchez2019hamiltonian}. Further, the acceleration is computed using Eq.~\ref{eq:ham}.

$\bullet \textbf{\hgnn:}$ \hgnn refers to Hamiltonian graph neural network, where the structure of the Hamiltonian of a system is exploited to decouple it into potential and kinetic energies~\cite{bhattoo2022learning,sanchez2019hamiltonian,greydanus2019hamiltonian}. Further, the potential energy is predicted using the \gnn and the diagonal mass matrix is trained as a learnable parameter. The acceleration is then predicted using Eq.~\ref{eq:ham}. Further details of \hgnn is provided in App.~\ref{app:hgnn}.
\looseness=-1

$\bullet$ \textbf{\chgn and \chgnn:} \chgn and \chgnn refer to \hgn and \hgnn with explicit constraints respectively (Eq.~\ref{eq:ham_constraint}-Eq.~\ref{eq:acc_ham}).  
\vspace{-0.1in}
\section{Benchmarking Evaluation}
\label{sec:experiments}
\begin{figure}[t]
    \centering
    \includegraphics[width=\columnwidth]{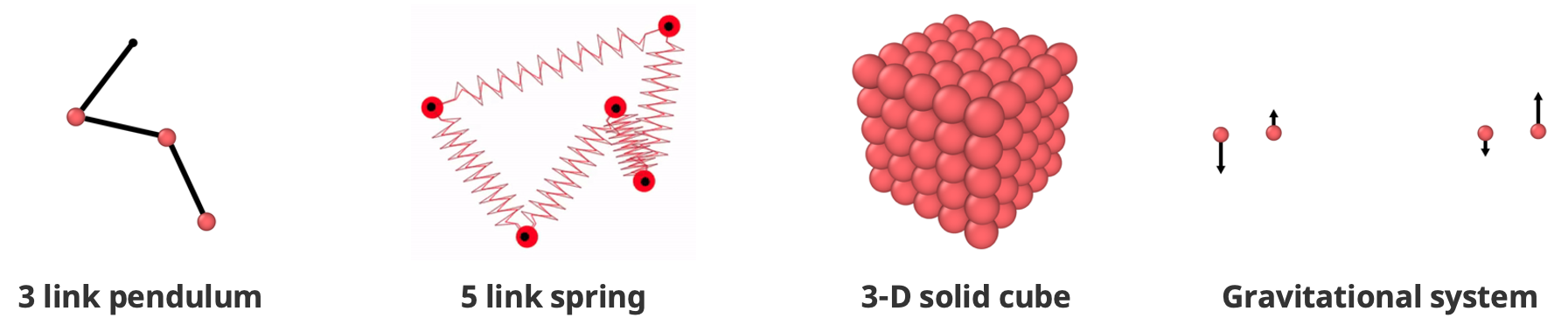}
    \caption{\rev{(From left to right) Visualizations of the systems considered, namely, 3-pendulum, 5-spring, elastically deformable 3D solid, and 4-body gravitational system.}}
    \label{fig:4systems}
    \vspace{-0.20in}
\end{figure}

\vspace{-0.10in}
In this section, we conduct in-depth empirical analysis of the architectures discussed in \S~\ref{sec:models}. \rev{To evaluate the models, we consider four systems (see Fig.~\ref{fig:4systems}), namely,} $n$-pendulums with $n = (3,4,5)$, $n$-springs where $n=(3,4,5)$, \rev{4-body gravitational system and an elastically deformable 3D solid discretized as particles.} 
All the simulations and training were carried out in the JAX environment~\cite{schoenholz2020jax,bradbury2020jax}. The graph architecture was developed using the jraph package~\cite{jraph2020github}. The experiments were conducted on a linux machine running Ubuntu 18.04 LTS with Intel Xeon processor and 128GB RAM. \rev{All codes and data used in the benchmarking study are available at \url{https://github.com/M3RG-IITD/benchmarking_graph} and \url{https://doi.org/10.5281/zenodo.7015041}, respectively.}

\vspace{-0.1in}
\subsection{Data Generation and Training}
\label{sec:datasets}
\vspace{-0.1in}

\textbf{Data generation.} The training data is generated by forward simulation using the known kinetic and potential energies of the systems employing physics-based equations. The timestep used for the forward simulation of the pendulum system is $10^{-5}s$ with the data collected every 1000 timesteps. For the spring system, the timestep is $10^{-3}s$ with the data collected every 100 timesteps. Integration of the equations of motion is performed using the velocity-Verlet algorithm. In both systems, 100 datapoints are collected from each trajectory and 100 such trajectories based on random initial conditions are used to generate ground truth. This resulted in a total of 10,000 training datapoints for each system. Further details of the experimental systems used for the data generation are provided in the App.~\ref{app:experiments}.
\looseness=-1

\textbf{Trajectory prediction and training.} Based on the predicted $\xddot$, the positions and velocities are predicted using the \textit{velocity Verlet} integration. The loss function is computed by using the predicted and actual accelerations at timesteps $2, 3,\ldots,\mathcal{T}$ in a trajectory $\mathbb{T}$, which is then back-propagated to train the \gnn{s}. Specifically, the loss function is as follows.
\begin{equation}
    \label{eq:lossfunction}
  \mathcal{L}= \frac{1}{n}\left(\sum_{i=1}^n \left(\ddot{x}_i^{\mathbb{T},t}-\left(\hat{\ddot{x}}_i^{\mathbb{T},t}\right)\right)^2\right)
\end{equation}
 Here, $\left(\hat{\ddot{x}}_i^{\mathbb{T},t}\right)$ is the predicted acceleration for the $i^{th}$ particle in trajectory $\mathbb{T}$ at time $t$ and $\ddot{x}_i^{\mathbb{T},t}$ is the true acceleration. $\mathbb{T}$ denotes a trajectory from $\mathfrak{T}$, the set of training trajectories. Note that the accelerations are computed directly from the ground truth trajectory using the Verlet algorithm as:
 \begin{equation}
\ddot{x}_i(t)=\frac{1}{(\Delta t)^2}[x_i(t+\Delta t)+x_i(t-\Delta t)-2x_i(t)]
\end{equation}
Since the integration of the equations of motion for the predicted trajectory is also performed using the same algorithm as: $x_i(t+\Delta t)=2x_i(t)-x_i(t-\Delta t)+\ddot{x}_i(\Delta t)^2$, this method is equivalent to training from trajectory (i.e, positions). 

We use 10000 data points generated from 100 trajectories to train all the models. This dataset is divided randomly in 75:25 ratio as training and validation set. All models were trained till the decrease in loss saturates to less than 0.001 over 100 epochs. The model performance is evaluated on a test set containing $100$ forward trajectories of $1s$ in the case of pendulum and $10s$ in the case of spring. Note that this trajectory is $\approx 4$-$5$ orders of magnitude larger than the training trajectories from which the training data has been sampled. The dynamics of $n$-body system is known to be chaotic for $n \geq 2$. Hence, all the results are averaged over trajectories generated from 100 different initial conditions.
\looseness=-1

\rev{The default hyper-parameter values are listed in App.~\ref{app:hyper}. In addition, we have performed extensive hyper-parameter search to measure their effect on the architectures, the details of which are provided in App.~\ref{app:hypersearch}.}
\looseness=-1

\vspace{-0.10in}
\subsection{Evaluation Metric} 
Following the work of~\cite{lnn1}, we evaluate  performance by computing the relative error in three different quantities as detailed below.
\textbf{(1) Rollout Error:} Relative error in  the trajectory, known as the \textit{rollout error}, is given by 
$RE(t)=\frac{||\hat{\x}(t)-\x(t)||_2}{||\hat{\x}(t)||_2+||\x(t)||_2}$.
\textbf{(2) Energy violation:} The error in energy violation is given by
$EE(t)=\frac{||\hat{H}-H||_2}{(||\hat{H}||_2+||H||_2)}$.
\textbf{(3) Momentum conservation:} The relative error in momentum conservation is
$ME(t)=\frac{||\hat{\mathcal{M}}-\mathcal{M}||_2}{||\hat{\mathcal{M}}||_2+||\mathcal{M}||_2}
\label{eq:mce}
$
Note that all the variables with a hat, for example $\hat{\x}$, represent the predicted values based on the trained model and the variables without hat, $\x$, represent the ground truth. To summarize the performance over a trajectory, following previous works~\cite{lnn1}, we use the \textit{geometric mean} of relative error of each of the quantities above since the error compounds with time.

\vspace{-0.1in}
\subsection{Results on Pendulum and Spring systems}
\vspace{-0.1in}
For springs and pendulums, the models are trained on $5$-spring, and $3$-pendulum systems. The trained models are evaluated on unseen system sizes to evaluate their performance on \textit{zero-shot} generalizability. To benchmark generalization ability to even larger \textit{unseen} systems, we simulate $5$-, $10$-, $20$-pendulum system and $5$-, $20$-, $50$-spring systems. 
 To compare the performance efficiently, we group the systems without and with explicit constraints separately.
\vspace{-0.1in}
\subsubsection{Trajectory error and energy conservation}
\begin{figure}
\vspace{-0.2in}
    \centering
    \begin{subfigure}{0.48\textwidth}
        \centering
        \includegraphics[width=0.99\textwidth]{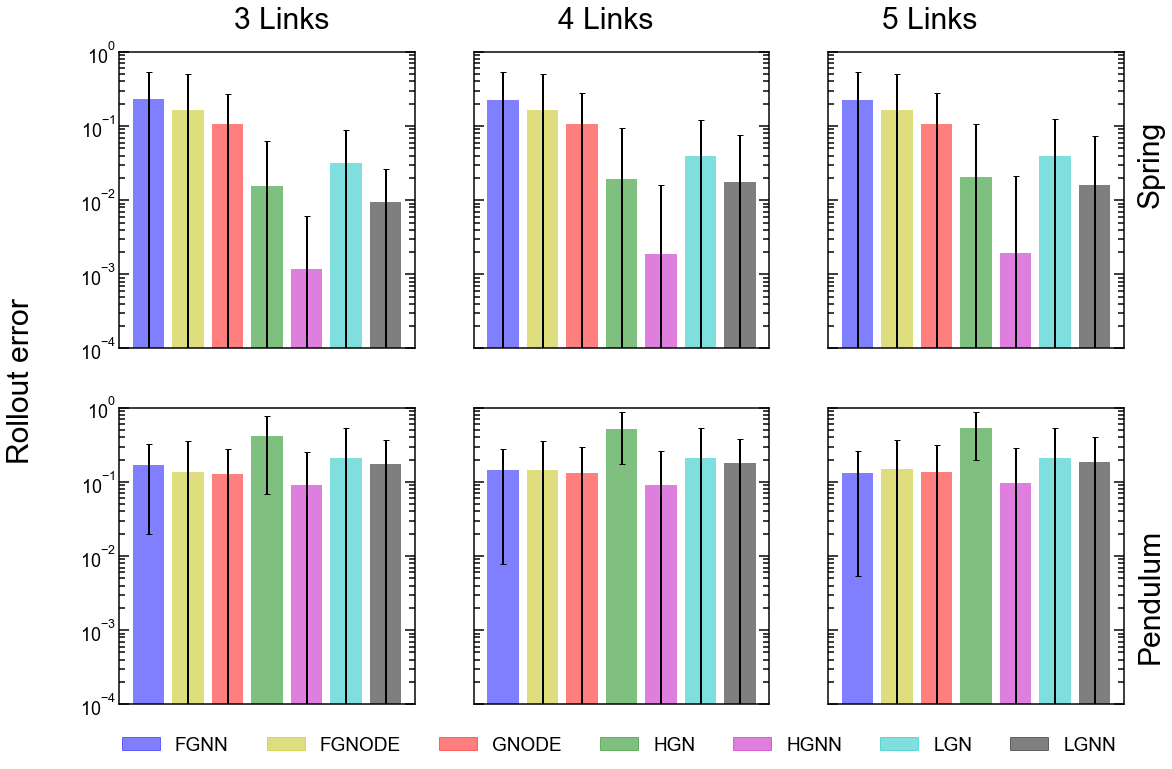} 
    \end{subfigure}\hfill
    \begin{subfigure}{0.48\textwidth}
        \centering
        \includegraphics[width=0.99\textwidth]{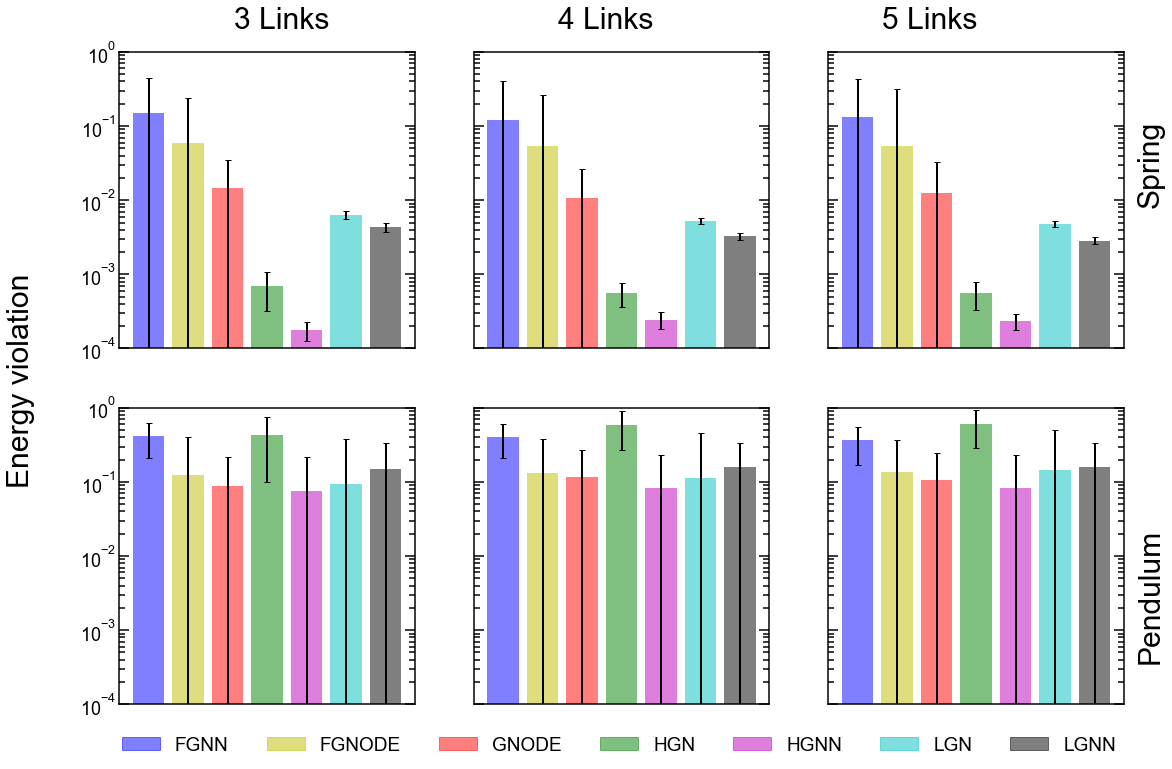} 
    \end{subfigure}
    \caption{Geometric mean of rollout error and energy error for 3-, 4-, 5-spring and 3-,4-,5-pendulum systems without constraints for \lgnn, \rev{\lgn}, \hgn, \hgnn, \gnode, \fgnode and \rev{\fgnn}. The error bar represents the 95\% confidence interval over 100 trajectories generated from random initial conditions.}
    \label{fig:unconstr}
    \vspace{-0.10in}
\end{figure}

\begin{figure}
    \centering
    \vspace{-0.10in}
    \begin{subfigure}{0.49\textwidth}
        \centering
        \includegraphics[width=0.99\textwidth]{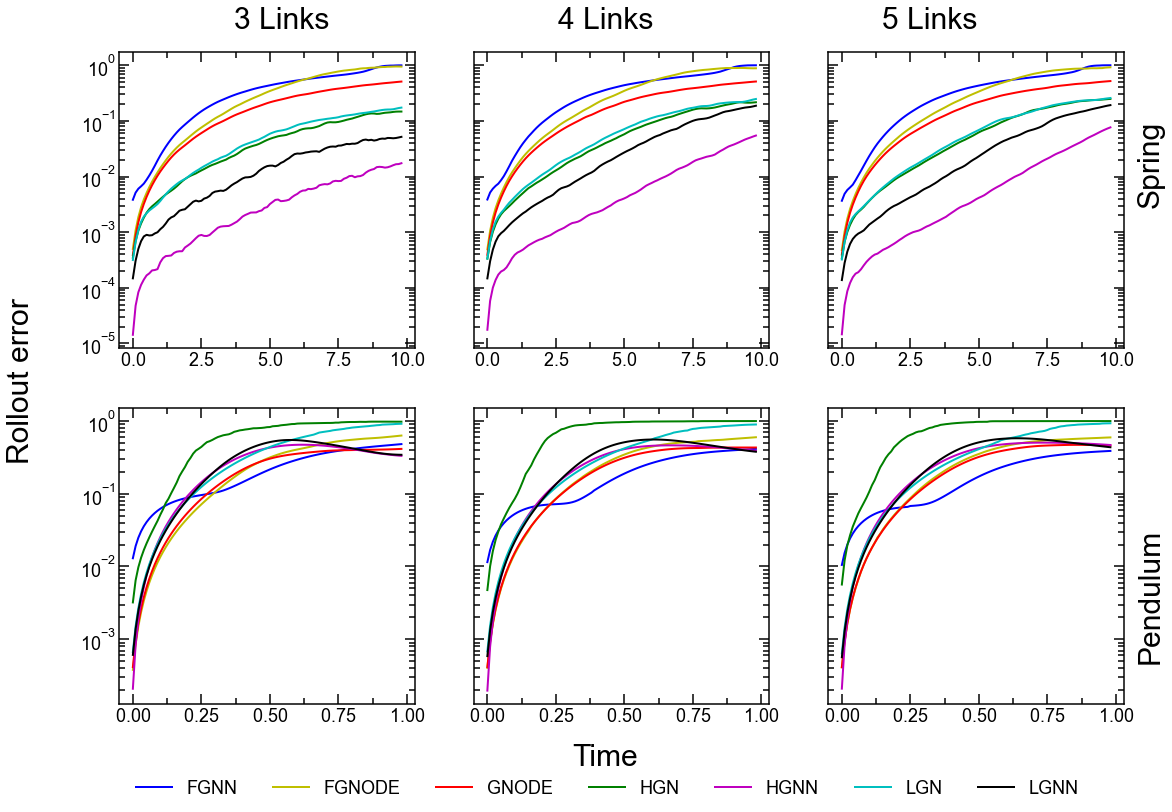} 
    \end{subfigure}\hfill
    \begin{subfigure}{0.49\textwidth}
        \centering
        \includegraphics[width=0.99\textwidth]{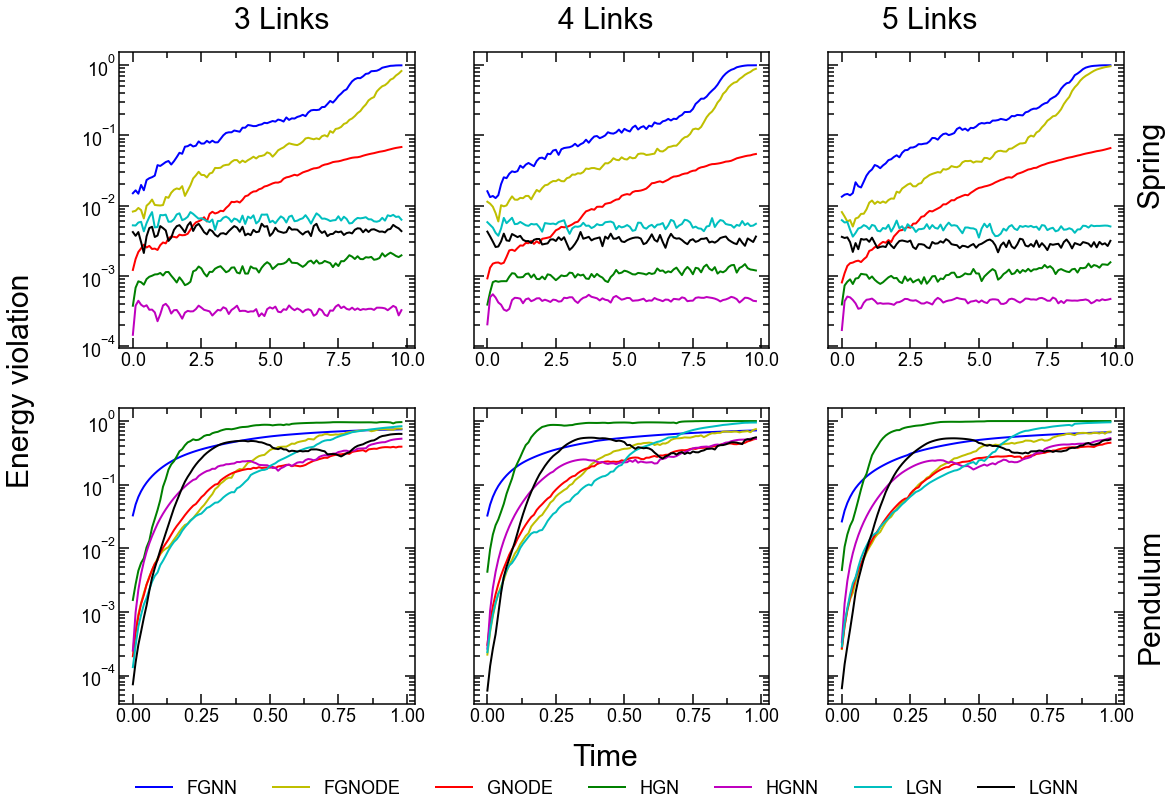} 
    \end{subfigure}
    \caption{ Rollout error and energy error for 3-, 4-, 5-spring and 3-, 4-, 5-pendulum systems without constraints with respect to time for \lgnn, \rev{\lgn}, \hgn, \hgnn, \gnode, \fgnode and \rev{\fgnn}. The curve represents the average over 100 trajectories generated from random initial conditions.}
    \label{fig:traj_unconstr}
    \vspace{-0.20in}
\end{figure}

\textbf{Systems without explicit constraints.} Figure~\ref{fig:unconstr} shows the geometric mean of rollout and energy error of \lgnn, \lgn, \hgn, \hgnn, \gnode, \fgnode and \fgnn for 3-, 4-, 5-spring and pendulum systems. Fig.~\ref{fig:traj_unconstr} shows the evolution of these errors with respect to time. Note that all the systems are trained on 3-pendulum or 5-spring systems alone and performance is evaluated by forward simulation on other systems. The error bar represents the 95\% confidence interval obtained based on 100 trajectories generated from different initial conditions. First, we analyze the response on the spring systems (see Fig.~\ref{fig:unconstr} top row). We observe that the \hgnn exhibits minimum error. This is followed by \lgnn and \hgn both of which comparable error but higher than that of \hgnn. These systems are followed by \gnode. \rev{Finally, \fgnn exhibits the maximum error. This result is not surprising since \fgnn is not physics-informed and learns directly from data.}

Now, we focus on the pendulum systems (see Fig.~\ref{fig:unconstr} bottom row). Interestingly, in this case, we observe that  \hgn exhibits a slightly higher error. It is worth noting that the dynamics of spring system is primarily governed by the internal interactions between the balls connected by the springs. In contrast, in pendulum systems, the dynamics is primarily governed by the external gravitational field and the connections simply serve as constraints. These results suggest that in systems, where the dynamics is governed primarily by the internal interactions, the architecture of the graph plays a major role with \hgnn exhibiting the best performance for the spring systems. \rev{\fgnn continues to perform poorly in pendulum systems, particularly in energy violation. }

In summary, two key observations emerge from these experiments. First, \hgnn consistently provides the lowest error across all setup, with it being more pronounced on spring systems, potentially due to its first-order nature. Second, the decoupled architectures are consistently better (i.e., \gnode and \hgnn produces lower error than \fgnode and \hgn respectively), which could be attributed to the ability of \gnn{s} to learn the parametric masses and potential energy functions, independently and uniquely. This observation is consistent with previous works on MLPs where decoupling is found to have significantly improved the performance of \hnn and \lnn~\cite{zhong2021benchmarking,lnn1,lnn2}.

\begin{figure}
    \centering
    \begin{subfigure}{0.49\textwidth}
        \centering
        \includegraphics[width=0.99\textwidth]{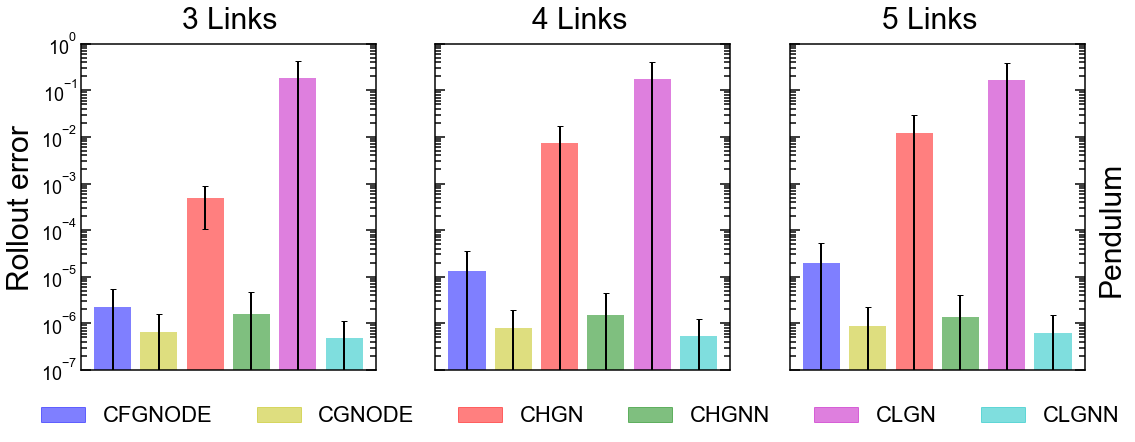} 
    \end{subfigure}\hfill
    \begin{subfigure}{0.49\textwidth}
        \centering
        \includegraphics[width=0.99\textwidth]{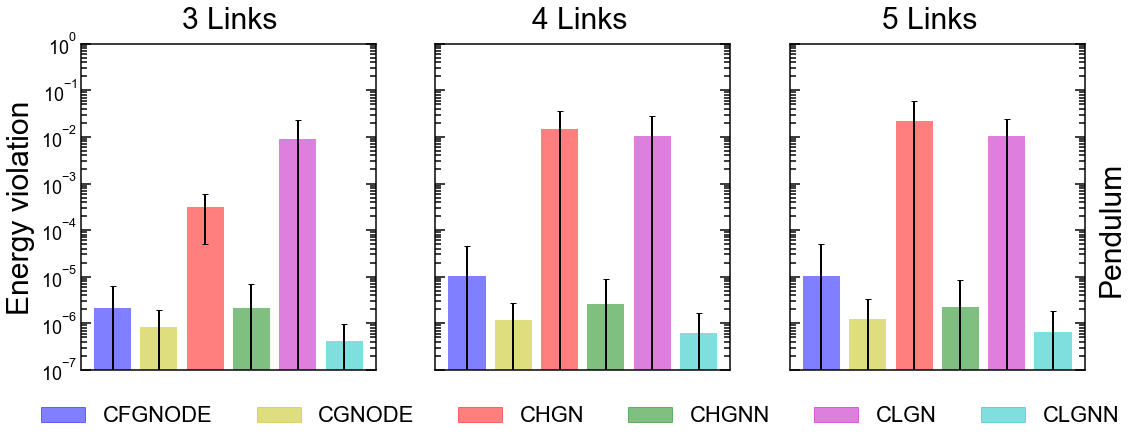} 
    \end{subfigure}
    \caption{Geometric mean of rollout error and energy error for 3-,4-,5-pendulum systems with constraints for \clgnn, \rev{\clgn}, \chgn, \chgnn, \cgnode, and \cfgnode. The error bar represents the 95\% confidence interval over 100 trajectories generated from random initial conditions.}
    \label{fig:constr}
\end{figure}

\begin{figure}[]
\vspace{-0.10in}
    \centering
    \begin{subfigure}{0.49\textwidth}
        \centering
        \includegraphics[width=0.99\textwidth]{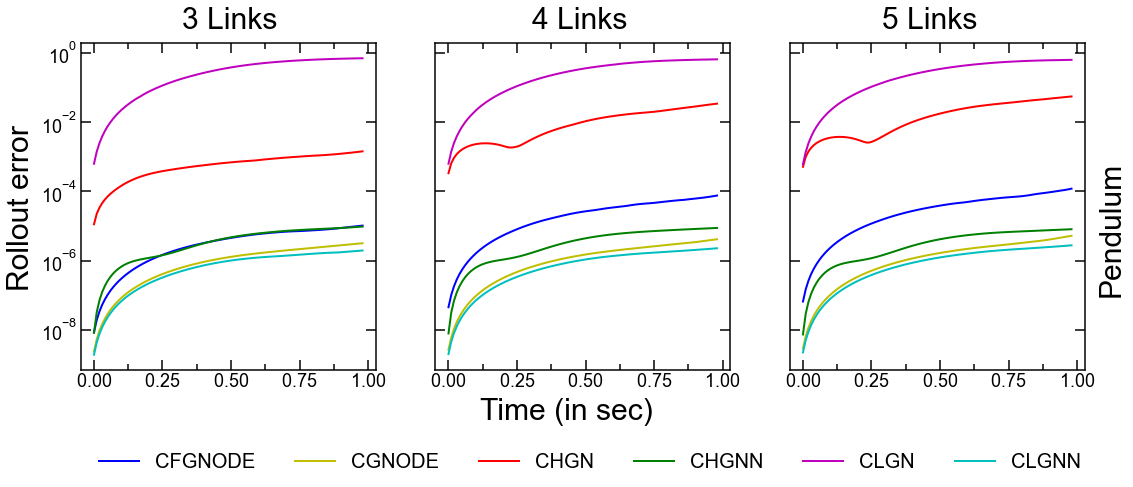} 
    \end{subfigure}\hfill
    \begin{subfigure}{0.49\textwidth}
        \centering
        \includegraphics[width=0.99\textwidth]{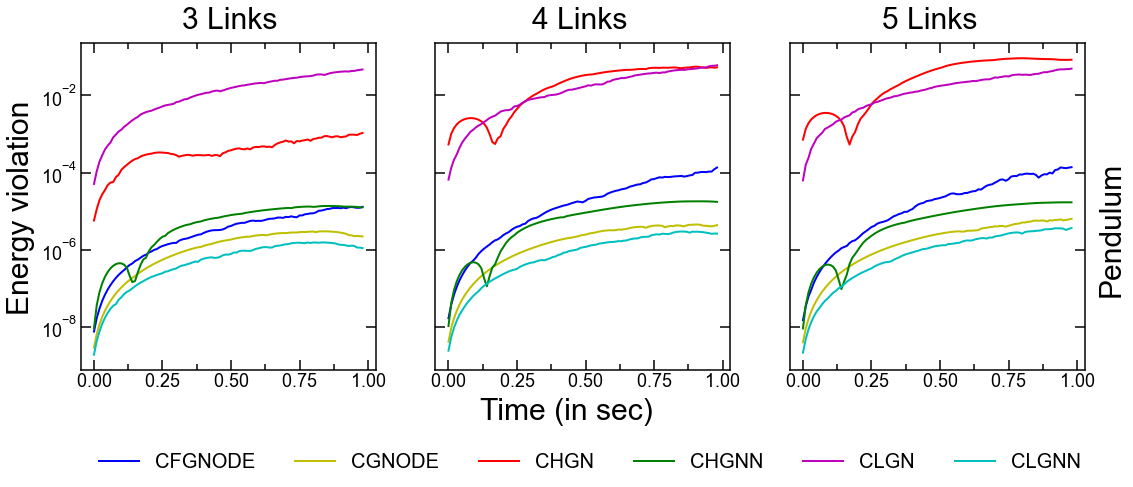}
    \end{subfigure}
    \caption{Rollout error and energy error for 3-,4-,5-pendulum systems with constraints with respect to time for \clgnn, \rev{\clgn}, \chgn, \chgnn, \cgnode, and \cfgnode. The curve represents the average over 100 trajectories generated from random initial conditions.}
    \label{fig:traj_constr}
    \vspace{-0.20in}
\end{figure}

\textbf{Systems with explicit constraints.} Figure~\ref{fig:constr} shows the performance of the constrained architectures for pendulum systems. Figure~\ref{fig:traj_constr} shows the time evolution of energy and rollout error. Note that in the spring system there are no explicit constraints. Hence, these approaches will be equivalent to learning without constraints in the spring systems as the constraints terms in the equations vanish.  We observe that the error in systems with explicit constraints are significantly lower than that without explicit constraints. Similar behavior was observed for fully-connected architectures earlier~\cite{lnn1}. 

\rev{We also observe that  \clgn exhibits the maximum error on average, followed closely by \chgn. The poor performance of \clgn is not surprising. In pendulum systems, the potential energy of the system depends on the position of the bob. However, in \lgn or \hgn family of architectures, the actual position of the particle is not given as an input to the graph, rather the edge distance is given as an input. Thus, it is not possible for an \lgn or \hgn to learn the dynamics of the pendulum system. To address this, we train \clgn and \chgn by giving the position of the bob explicitly as a node input feature. However, despite providing this input, we observe that the final model obtained after training has high loss in comparison to  other models.}


\vspace{-0.10in}
\subsubsection{Momentum conservation}
\vspace{-0.10in}
\begin{wrapfigure}{r}{0.60\textwidth}
\vspace{-0.20in}
    \centering
        \includegraphics[width=0.58\textwidth]{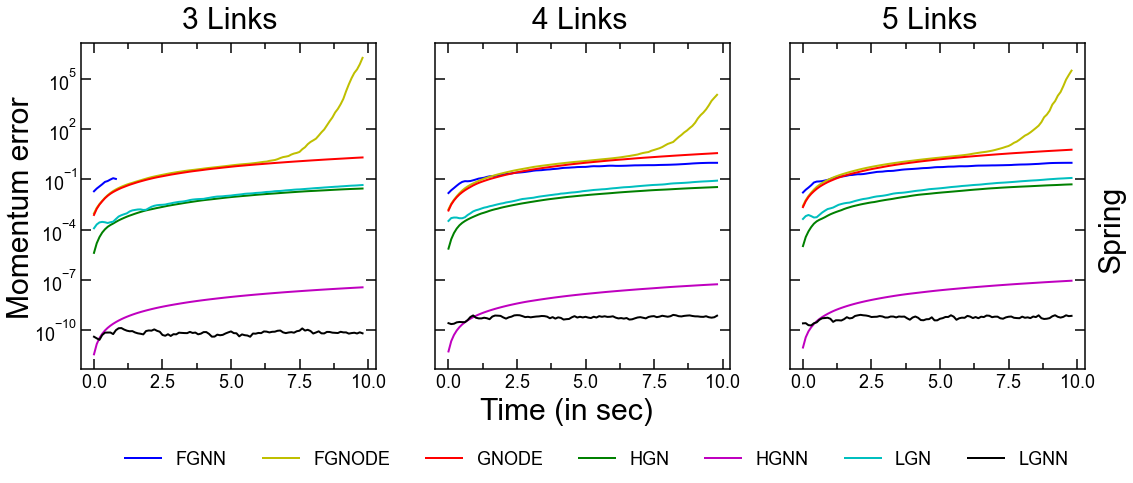} 
    \vspace{-0.10in}
    \caption{Momentum error for spring systems 3-, 4-, and 5- spring systems for \lgnn, \lgn, \hgn, \hgnn, \gnode, \fgnode and \fgnn. The curve represents the average over 100 trajectories generated from random initial conditions.}
    \label{fig:mom_cons}
    \vspace{-0.10in}
\end{wrapfigure}

While most of the previous studies have focused on energy and rollout error, very few studies have analyzed the momentum error in the trajectory predicted by these systems. It should be noted that conservation of momentum is one of the fundamental laws (as much as the energy conservation), which physical systems are expected to follow. While violation of energy conservation results in spurious generation or dissipation of energy in the system, violation of momentum conservation results in such effects in the total force on a particle in the system. This, in turn, affects the force equilibrium of a system. Figure~\ref{fig:mom_cons} shows the evolution of momentum error in \fgnn, \fgnode, \gnode, \hgn, \hgnn, \lgn and \lgnn for 3-, 4-, and 5-spring systems. Interestingly, we observe that \lgnn exhibits least momentum error with a stable value over time. This suggests that the momentum error in the \lgnn is saturated and does not diverge over time. Although \hgnn exhibits comparable errors, we observe that the value increases with respect to time suggesting a potential divergence at long time scales. This is followed by \hgn, which exhibits a momentum error that $\sim$ 4 orders of magnitude larger. While \fgnode and \gnode exhibits comparable momentum error at smaller length scales of trajectory, \fgnode diverges significantly faster that \gnode, leading to large unrealistic errors. \rev{The improved performed of \lgnn and \hgnn could be attributed to the decoupling of kinetic and potential energies, and the computation of the potential energy at the edge level, which enforces the momentum conservation indirectly due to the translational symmetry.} These results reaffirm that, in interacting systems, the graph architecture and the nature of the inductive bias (\node vs \lnn vs \hnn) play crucial roles in governing the stability of the system.

\vspace{-0.10in}
\subsubsection{Generalizability to unseen system sizes}
\label{sec:generalizability}
\vspace{-0.10in}
\begin{figure}
\vspace{-0.20in}
    \centering
    \begin{subfigure}{0.49\textwidth}
        \centering
        \includegraphics[width=0.99\textwidth]{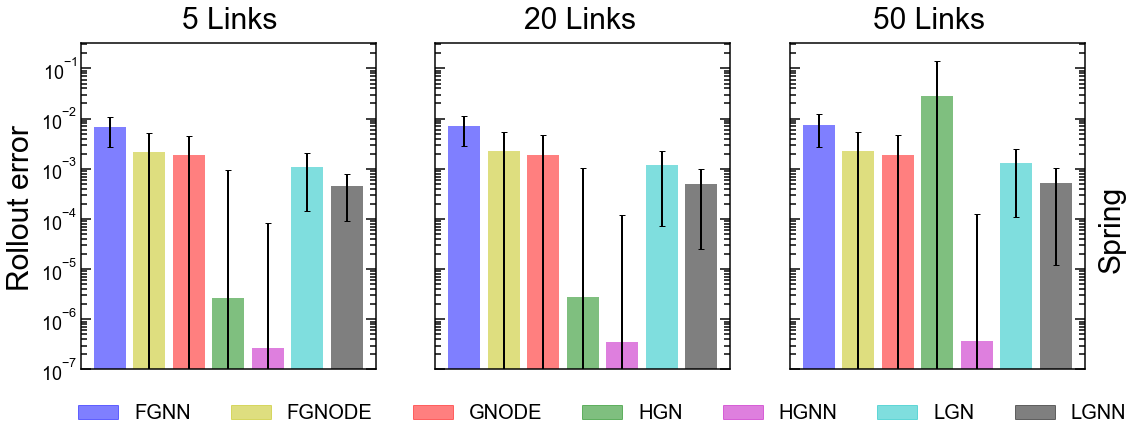} 
    \end{subfigure}
    \begin{subfigure}{0.49\textwidth}
        \centering
        \includegraphics[width=0.99\textwidth]{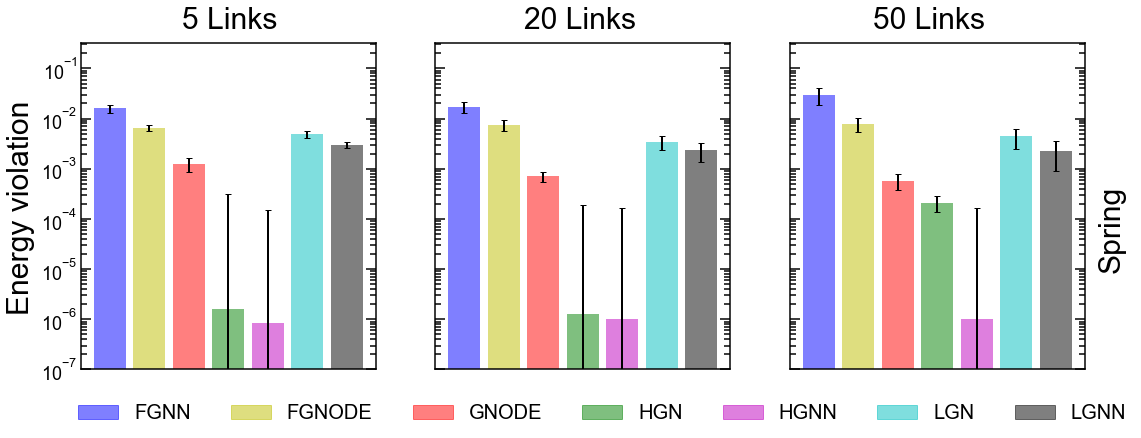} 
    \end{subfigure}
    \begin{subfigure}{0.49\textwidth}
        \centering
        \includegraphics[width=0.99\textwidth]{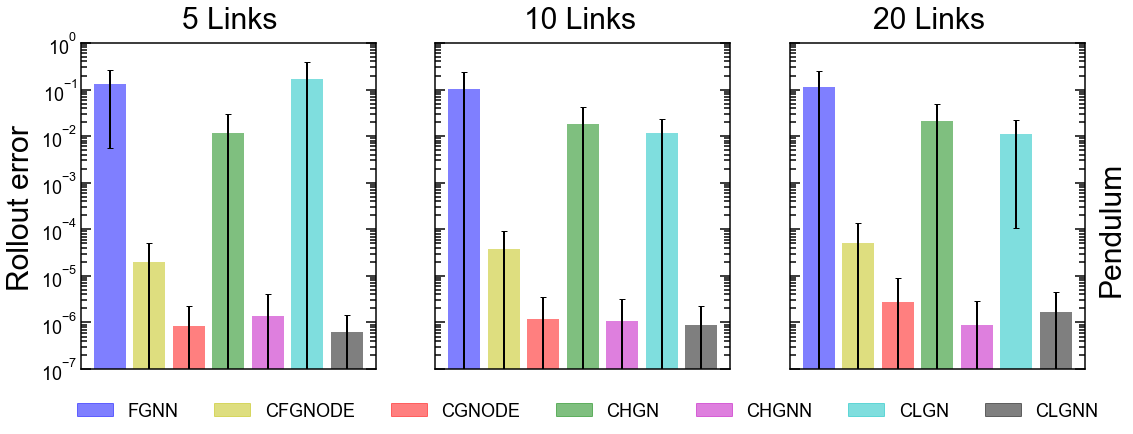} 
    \end{subfigure}
    \begin{subfigure}{0.49\textwidth}
        \centering
        \includegraphics[width=0.99\textwidth]{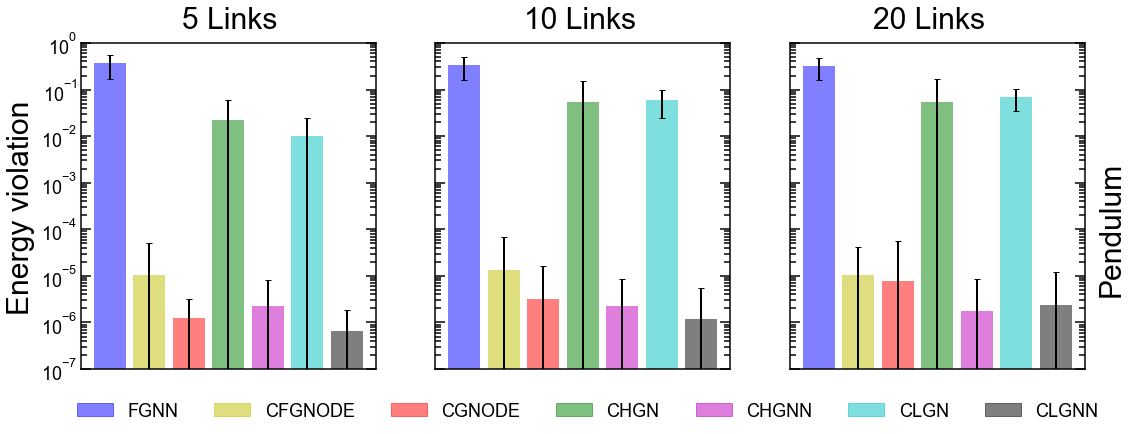} 
    \end{subfigure}
    
    \caption{Geometric mean of rollout and energy error for zero-shot generalizability to unseen system sizes for pendulum systems with \fgnn, \clgn, \clgnn, \chgn, \chgnn, \cgnode, and \cfgnode and spring systems with \fgnn, \lgn, \lgnn, \hgn, \hgnn, \gnode, \fgnode. The error bar represents the 95\% confidence interval over 100 trajectories generated from random initial conditions.}
    \label{fig:generalizability}
    \vspace{-0.20in}
\end{figure}

Now, we analyze the ability of the \gnn{s} to generalize to unseen system sizes. To this extent, we evaluate the performance of the constrained systems for 5-, 10-, and 20-link pendulums and unconstrained systems for 5-, 20-, and 50-link springs. Note that the constrained systems are considered for the pendulum as they exhibit superior performance in comparison to the unconstrained versions. Figure~\ref{fig:generalizability} shows the geometric mean of rollout and energy error for these systems. First, we observe that all the graph-based systems exhibit zero-shot generalizability to systems that are almost an order of magnitude larger than the training data with similar errors in both energy and rollout. Further, we observe that the trend in the error for the systems for pendulum and spring systems remain consistent with increasing system sizes. This suggests that there is no additional error added to any of the models considered due to zero-shot generalization. This confirms the superiority of physics-informed \gnn{s} over the traditional MLPs to generate large-scale realistic systems after learning on a significantly smaller system. 
\vspace{-0.10in}
\subsection{\rev{Gravitational and 3D solid systems}}
\begin{figure}[t]
    \centering
    \begin{subfigure}{0.22\textwidth}
        \centering
        \includegraphics[width=0.99\textwidth]{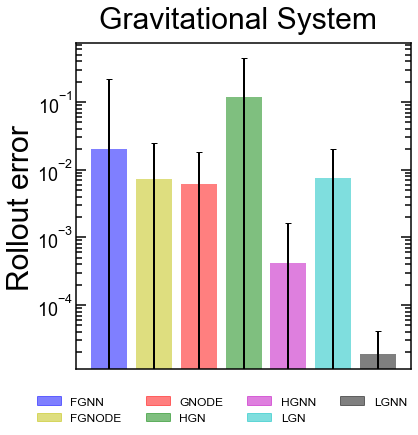} 
    \end{subfigure}\hfill
    \begin{subfigure}{0.22\textwidth}
        \centering
        \includegraphics[width=0.99\textwidth]{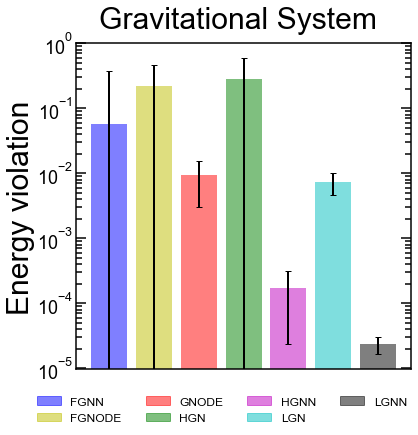} 
    \end{subfigure}
    \begin{subfigure}{0.22\textwidth}
        \centering
        \includegraphics[width=0.99\textwidth]{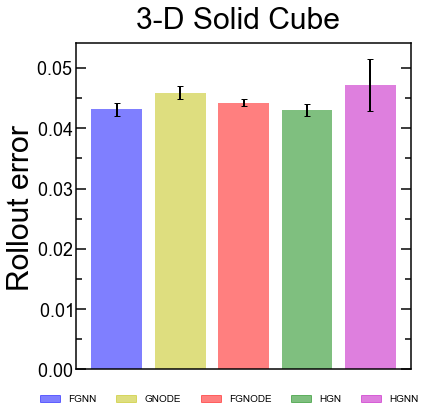} 
    \end{subfigure}\hfill
    \begin{subfigure}{0.22\textwidth}
        \centering
        \includegraphics[width=0.99\textwidth]{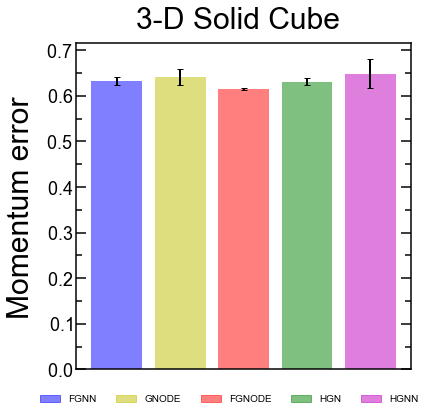} 
    \end{subfigure}
\caption{\rev{(left to right) Geometric mean of rollout error, energy error for 4-body gravitational system for \lgnn, \lgn \hgn, \hgnn, \gnode, \fgnode and \fgnn. Geometric mean of rollout error and momentum error for rigid body system for \hgn, \hgnn, \gnode, \fgnode and \fgnn. The error bar represents the 95\% confidence interval over 100 trajectories generated from random initial conditions.} \label{fig:rigid_bar}}
\vspace{-0.40in}
\end{figure}

\vspace{-0.10in}
\rev{To test the performance of the models on more complex systems, we consider a 4-body gravitational system and a 3D solid system (see Fig.~\ref{fig:4systems}). For gravitational system, a stable four body configuration interacting with each other through the gravitational law and having initial velocities such that they rotate with respect to a common centre is considered (see: App.~\ref{app:experiments}). Figure~\ref{fig:rigid_bar} shows the time evolution of energy and rollout error for these systems. We observe that \lgnn exhibits the lowest error followed by \hgnn. This is followed by \lgn and \gnode{} which exhibits comparable errors.}

\rev{Now, we evaluate the performance on the elastic deformation of a 3D solid cube. For this, a $5 \times 5 \times 5$ solid cube, discretized into $125$ particles, is used for simulating the ground truth in the peridynamics framework. The system is compressed isotropically and then released to simulate the dynamics of the solid, that is, the contraction and expansion in 3D as a function of time. Figure~\ref{fig:rigid_bar} shows the rollout and momentum error for this system (see Fig.~\ref{fig:rigid_traj} for time evolution). Interestingly, we find that the rollout and momentum error for all the models are comparable. This could be attributed to the nature of deformation of the structure, which exhibits only small displacements within the elastic regime. Note that this dynamics is much simpler and non-chaotic in nature. Thus, all the models are able to learn and infer the dynamics with similar accuracy. It may be concluded from this experiment that for dynamics that is not chaotic and relatively small in terms of magnitude, physics-informed \gnns may not have any additional advantage over the data-driven ones.}
\section{Concluding Insights}
\label{sec:conclusion}
In this work, we benchmark the performance of \rev{thirteen} different physics-informed \gnn{s} for simulating interacting physical systems such as springs and pendulums. The key insights drawn from our study are summarized below:
\color{black}
\begin{itemize}
    \item \textbf{Architecture matters:} We show the nature of the inductive bias provided by the ODE, Hamiltonian and Lagrangian equations lead to different performances for the same graph architecture, although these equations are, in-principle, equivalent~\cite{zhong2021benchmarking}. Specifically, the ODE and Lagrangian formulations result in a second-order differential equation, while the Hamiltonian formulation results in a first-order differential equation. 
    
    \item \textbf{Physics-informed \gnns are better:} All of the physics-informed \gnns exhibit better performance than \fgnn, that is learning purely from data except in the case of small displacements that are non-chaotic in nature (elastic deformation of 3D solid). The inductive bias provided for learning and inference by Lagrangian, Hamiltonian, and Newtonian mechanics to \gnns are clearly distinct, although the formulations are equivalent in principle.
    
      \item \textbf{Constraints help:} Incorporating constraints explicitly as an inductive bias significantly simplifies the learning (see Fig.~\ref{fig:dataefficiency}) and enhances the performance of all the graph-based models.
      
      \item \textbf{Kinetic and potential energies should be decoupled:} We observe that exploiting the fact that the kinetic and potential energies can be decoupled (in Cartesian coordinates) leads to improved performance in all the models---\gnode better than \fgnode, \hgnn better than \hgn, and \lgnn better than \lgn for most systems. This decoupling exhibits a significantly stronger enforcement of the momentum conservation error. This could be attributed to the fact that, when the potential and kinetic energies are decoupled, additional spurious cross-terms involving both $\x$ and $\xdot$ may not be learned by the model, which leads to lower error in the governing equation and the momentum error.
      \looseness=-1
      
      \item \textbf{Zero-shot generalizability:}  We show that all the graph-based models generalize to larger systems with comparable error as the smaller systems. Thus, a graph-based simulator with low error in small systems shall also exhibit similar error in significantly larger systems---thanks to their inductive bias.

    \item {\bf The \lgn family is slow to train: } Due to the double derivative present in the Lagrangian, the \lgn family exhibits highest training and inference times in comparison to other models (See App.~\ref{app:time}). These slow running times do not provide any benefit on accuracy. Hence, \gnode and \hgn provide better efficiency-efficacy balance while making a \gnn physics-informed.
      
\end{itemize} 
\textbf{Limitations and outlook.} 
\rev{Our results clearly highlight the importance of constraints. Current physics-informed \gnns require these constraints to be explicitly incorporated. An important future direction of research would therefore be to \textit{learn} the constraints directly from the trajectory. In addition, incorporating deformations, contacts, and other realistic phenomena remains to be explored in physics-informed graph architecture, although similar approaches have been employed in fully connected MLPs~\cite{zhong2020dissipative,zhong2021extending}.} \rev{Finally, we observe that there is a 10-fold increase in error in the presence of noise in the training data (see App.~\ref{app:noise}). This behavior points towards the need to build architectures that are more robust to noise.}
\color{black}

\section*{Acknowledgments}
The authors thank IIT Delhi HPC facility for computational and storage resources.

\bibliographystyle{apalike}
\bibliography{example_paper}

\maketitlesup
\appendix
\section{Expressing constraints}
\label{app:constraints}

\rev{A constraint on a system essentially restricts the motion of a system to a subspace among all the allowable paths. For instance, in the case of two particles with the coordinates $(x,y)$ and $(0,0)$ connected by an in-extensible rod, the constraint equation can be given as $(x^2 + y^2) =l^2$. Such constraints are known \textit{holonomic} constraints. However, a different set of constraints act in cases such as multi-fingered grasping, known as \textit{Pfaffian} constraints, where instead of positions, constraints are enforced on velocities. The generic form of a Pfaffian constraint is $A(\cx)\dot{\cx}= 0$. Note that any holonomic constraint can also be written in the form of a Pfaffian constraint by differentiating the original form. For instance, the constraint equation for two particles mentioned earlier can be differentiated to obtain $\cx\dot{\cx}+\cy\dot{\cy}=0$, which is of the form $A(\cx)\dot{\cx}= 0$. For the sake of generality, here we adopt this form to express the constraints. More details on this can be found in the Section 1, Chapter 6 of Murray et al. (Murray, R.M., Li, Z. and Sastry, S.S., 2017. A mathematical introduction to robotic manipulation. CRC press.)}

\section{Experimental systems}
\label{app:experiments}
To simulate the ground truth, physics-based equations derived using Lagrangian mechanics is employed. The equations for $n$-pendulum and spring systems are given in detail below.
\subsection{$n$-Pendulum}
For an $n$-pendulum system, $n$-point masses, representing the bobs, are connected by rigid (non-deformable) bars. These bars, thus, impose a distance constraint between two point masses as 
\begin{equation}
\rev{||x_{i}-x_{i-1}||^2 = l_i^2}
\end{equation}
where, $l_i$ represents the length of the bar connecting the $(i-1)^{th}$ and $i^{th}$ mass. This constraint can be differentiated to write in the form of a \textit{Pfaffian} constraint as
\begin{equation}
    \rev{(x_i-x_{i-1})(\dot{x}_i-\dot{x}_{i-1})=0}
\end{equation}
Note that such constraint can be obtained for each of the $n$ masses considered to obtain the $A(q)$.

The Lagrangian of this system can be written as
\begin{equation}
    L=\sum_{i=1}^n \left(1/2m_i\dot{x_i}^\texttt{T}\dot{x_i}-m_igx^{(2)}_i\right)
\end{equation}
where $m_i$ represents the mass of the $i^{th}$ particle, $g$ represents the acceleration due to gravity in the $x^{(2)}$ direction and $x^{(2)}$ represents the position of the particle in the $x^{(2)}$
direction. 
\subsection{$n$-spring system}
Here, $n$-point masses are connected by elastic springs that deform linearly (elastically) with extension or compression. Note that similar to the pendulum setup, each mass $m_i$ is connected to two masses $m_{i-1}$ and $m_{i+1}$ through springs so that all the masses form a closed connection. The Lagrangian of this system is given by
\begin{equation}
    L=\sum_{i=1}^n 1/2m_i\dot{x_i}^\texttt{T}\dot{x_i}- \sum_{i=1}^n 1/2k(||x_{i-1}-x_{i}||-r_0)^2
\end{equation}
where $r_0$ and $k$ represent the undeformed length and the stiffness, respectively, of the spring.
\rev{\subsection{$n$-body gravitational system}
Here, $n$ point masses are in a gravitational field generated by the point masses themselves. The Lagrangian of this system is given by
\begin{equation}
    L=\sum_{i=1}^n 1/2m_i\dot{x_i}^\texttt{T}\dot{x_i}+ \sum_{i=1}^n\sum_{j=1,j\neq i}^n Gm_im_j/2(||x_{i}-x_{j}||)
\end{equation}
where $G$ represents the Gravitational constant.}
\rev{\subsection{Rigid-body system}}
\rev{Here, in a solid cube of $5 \times 5 \times 5$ size, the dynamics of an elastically deformable body is simulated. Specifically, a $5 \times 5 \times 5$ solid cube, discretized into $125$ particles, is used for simulating the ground truth. 3D solid system is simulated using the peridynamics framework. The system is compressed isotropically and then released to simulate the dynamics of the solid, that is, the contraction and expansion in 3D.}

\section{Graph Neural ODE (\gnode)}
\label{app:gnode}
\begin{figure}
\centering
\includegraphics[width=\columnwidth]{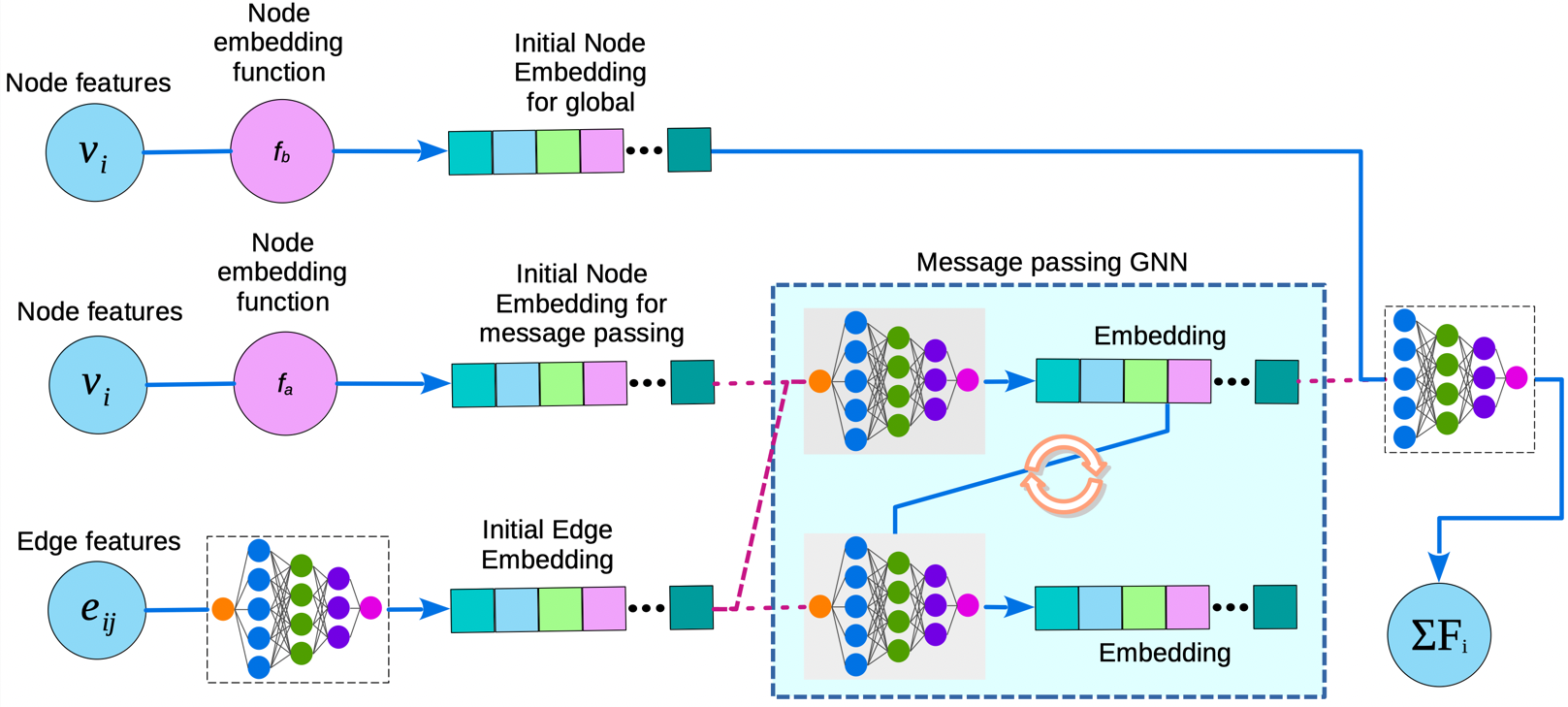}
\caption{\gnode architecture.}
\label{fig:gnode_arch}
\end{figure}

To learn the dynamical systems, \gnode{}s parameterize the dynamics $F(\x,\xdot,t)$ using a neural network to learn the approximate function $\hat{F}(\x_t,\xdot_t,t)$ by minimizing the loss between the predicted and actual trajectories, that is, $\mathcal{L} = ||\x_{t+1} - \hat{\x}_{t+1}||$. Thus, a \gnode{} essentially uses graph topology to learn the approximate dynamics $\hat{F}$ by training directly from the trajectory. Figure~\ref{fig:gnode_arch} shows the architecture of the \gnode, which is discussed in detail below.\\
{\bf Graph structure.} 
\label{sec:graph_cal}
First, an $n$-particle system is represented as a undirected graph $\mathcal{G}=\{\mathcal{V,E}\}$, where the nodes represent the particles and the edges represents the connections or interactions between them. For example, in pendulum or spring systems, the nodes correspond to bobs or balls, respectively, and the edges correspond to the bars or springs, respectively.\\
{\bf Input features.} Each node is characterized by the features of the particles, namely, the particle \textit{type} ($t$), \textit{position} ($x_i$), and \textit{velocity} ($\dot{x}_i$). The \textit{type} distinguishes particles of differing characteristics, for instance, balls or bobs with different masses. Further, each edge is represented by the edge features $w_{ij}=(x_i-x_j)$, which represents the relative displacement of the nodes connected by the given edge.\\
\textbf{Pre-Processing.} In the pre-processing layer, we construct a dense vector representation for each node $v_i$ and edge $e_{ij}$ using $\texttt{MLP}_{em}$ as:
\begin{alignat}{2}
    \ch^0_i &= \sqp(\MLP_{em}(\texttt{one-hot}(t_i),x_i,\dot{x}_i)) \label{eq:one-hot}\\
    \ch^0_{ij} &= \sqp(\MLP_{em}(w_{ij}))
\end{alignat}
$\sqp$ is an activation function. Note that the $\MLP_{em}$ corresponding to the node and edge embedding functions are parameterized with different weights. Here, for the sake of brevity, we simply mention them as $\MLP_{em}$.\\
\textbf{Acceleration prediction.} In many cases, internal forces in a system that govern the dynamics are closely dependent on the topology of the structure. To capture this information, we employ multiple layers of \textit{message-passing} between the nodes and edges. In the $l^{th}$ layer of message passing, the node embedding is updated as:
\begin{equation}
    \ch_i^{l+1} = \texttt{squareplus} \left( \ch_i^{l}+\sum_{j \in \mathcal{N}_i}\cW_{\CV}^l\cdot\left(\ch_j^l || \ch_{ij}^l\right) \right)
\end{equation}
where, $\mathcal{N}_i=\{v_j\in\CV\mid e_{ij}\in\CE \}$ are the neighbors of $v_i$. $\cW_{\CV}^{l}$ is a layer-specific learnable weight matrix.
$\ch_{ij}^l$ represents the embedding of incoming edge $e_{ij}$ on $v_i$ in the $l^{th}$ layer, which is computed as follows.
\begin{equation}
    \ch_{ij}^{l+1} = \texttt{squareplus} \left( \ch_{ij}^{l} + \cW_{\CE}^{l}\cdot\left(\ch_i^l || \ch_{j}^l\right) \right)
\end{equation}
Similar to $\cW_{\CV}^{l}$, $\cW_{\CE}^{l}$ is a layer-specific learnable weight matrix specific to the edge set. The message passing is performed over $L$ layers, where $L$ is a hyper-parameter. The final node and edge representations in the $L^{th}$ layer are denoted as $\ch_i^L$ and $\ch_{ij}^L$ respectively. 

In addition to the internal forces, there could be forces that are independent of the topology and depend only on the features of the particle, for example, gravitational force. To account for these, an additional node embedding that is not included in the message passing, namely, $\ch_i^{g}$ is concatenated with the final node representation after message passing as $\cz_i=(\ch_i^L||\ch_i^g)$. Finally, the acceleration of the particle $\ddot q_i$ is predicted as:
\begin{equation}
\ddot x_i=\texttt{squareplus}(\texttt{MLP}_{\CV}(\cz_i))
\end{equation}

Note that the major difference between \gnode and \fgnode, in addition to the other parametric and architectural differences, is the inclusion of this additional global feature embedding in \gnode, which is absent in \fgnode. As seen earlier, inclusion of this additional embedding significantly improves the performance in cases where there are forces due to external fields such as gravity.

\section{Lagrangian Graph Neural Network (\lgnn)}
\label{app:lgnn}
\begin{figure}
\centering
\includegraphics[width=\columnwidth]{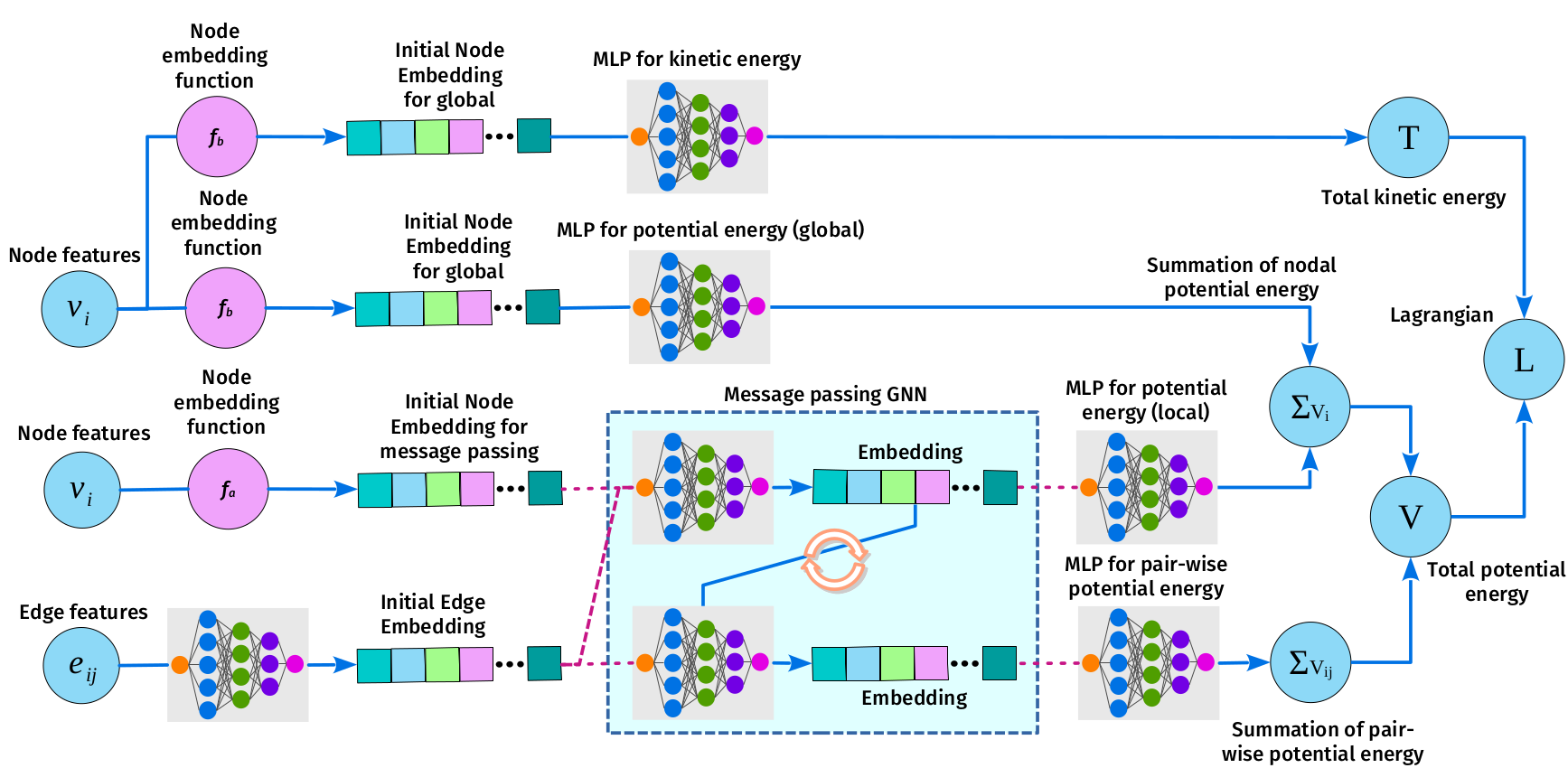}
\caption{\lgnn architecture.}
\label{fig:lgnn_arch}
\end{figure}

Figure~\ref{fig:lgnn_arch} shows the the neural network architecture of the \lgnn. The architecture directly predicts the Lagrangian of the system exploiting the topology of the system encoded as graph. Note that the graph structure, input features, pre-processing and the message passing leading to intermediate embedding of nodes and edges for \lgnn are developed exactly following the same as \gnode, detailed as per section \ref{sec:graph_cal}. Two major differences for \lgnn from \gnode is in the computation of the kinetic energy and the potential energy, which are absent in \gnode, as detailed below.





\textbf{Kinetic energy.} Since the system is comprised of $n$-point particles, the mass matrix becomes diagonal in Cartesian coordinates~\cite{lnn1}. Thus, the kinetic energy, $\tau_i$, of a point particle depends only on its velocity and mass. Here, we learn the mass of each particle based on the node embedding $\ch_i^0$ as 
\begin{equation}
    \tau_i = \texttt{squareplus}(\texttt{MLP}_{\tau}(\ch^0_i \parallel \dot{x}_i))
\end{equation}
The kinetic energy of the individual particles are summed up to compute the total kinetic energy as
\begin{equation}
    T = \sum_{u_i\in\mathcal{U}} \tau_i
\end{equation}

\textbf{Potential energy.} Potential energy of a system can have complex combination of absolute and topological features. For instance, a system such as a pendulum in a gravitational field have a simple potential energy function that does not depend on topology. On the contrary, for a system such as balls connected with spring, the potential energy depends on the connections and hence, the topology. Therefore, similar to \gnode, we use final node ($\ch^L_i$) and edge ($\ch^L_{ij}$) embedding from message passing (representing topology) and global node ($\ch_i^g$) embedding to calculate the potential energy of the system as
\begin{equation}
    V= \sum_{u_i\in\CU} v_i + \sum_{e_{ij}\in\CE} v_{ij}
\end{equation}
where $v_i =\texttt{squareplus}(\texttt{MLP}_{v_i}(\ch_i^g) + \texttt{squareplus}(\texttt{MLP}_{\texttt{mp},v_i}(\ch^L_i))$ represents the energy due to the attributes of the particle themselves, and $v_{ij} = \texttt{squareplus} (\texttt{MLP}_{v_{ij}}(\ch^L_{ij}))$ represents energy due to interactions (topology).

\textbf{Lagrangian.} 
Finally, the Lagrangian of the system is defined as $L = T - V$ where $T$ is total kinetic energy of system and $V$ is the total potential energy of the system. Finally, the acceleration is computed using the predicted Lagrangian employing the appropriate $EL$ equation.

\section{Hamiltonian Graph Neural Network (\hgnn)}
\label{app:hgnn}
\begin{figure}
\centering
\includegraphics[width=\columnwidth]{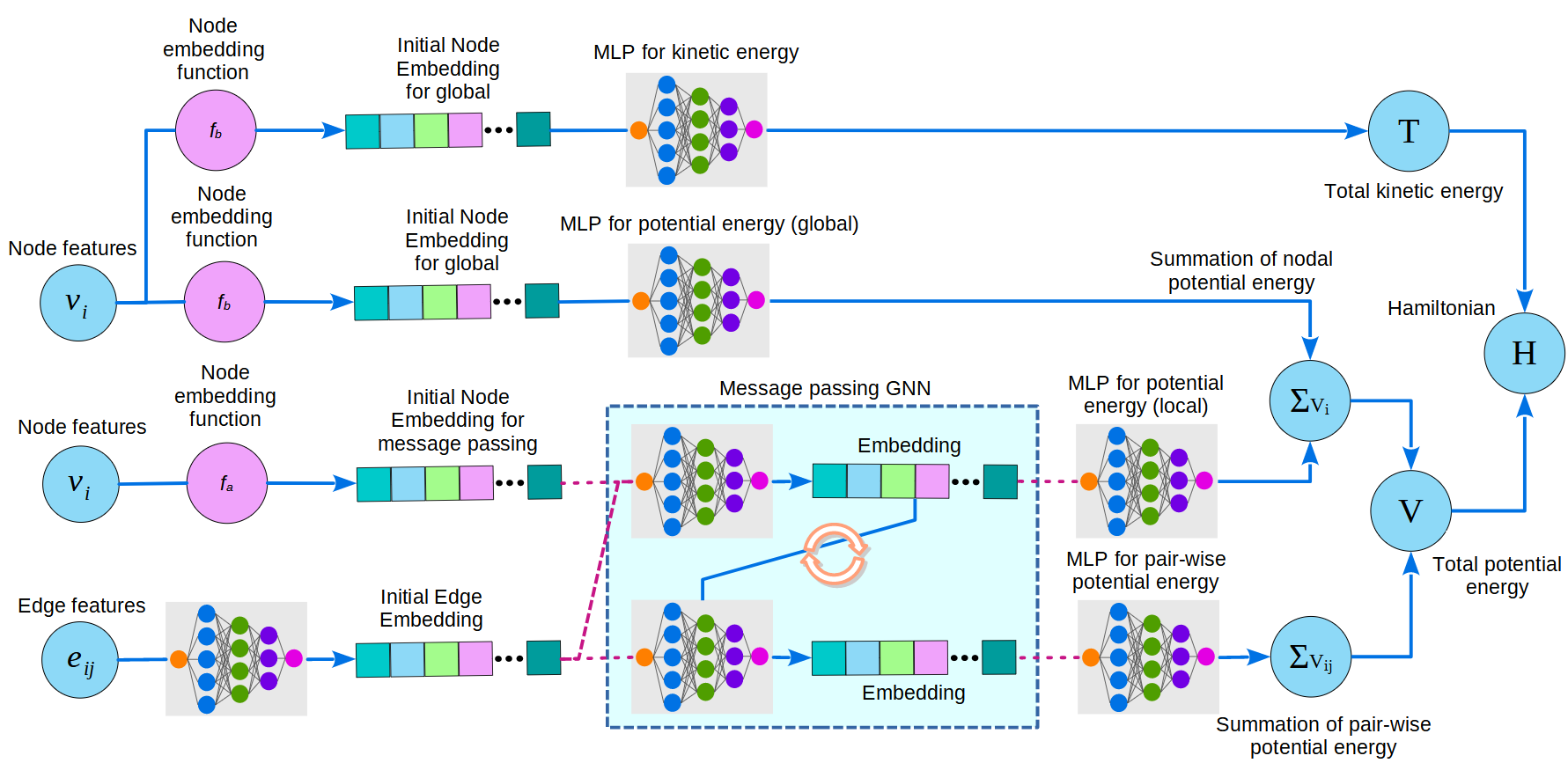}
\caption{\hgnn architecture.}
\label{fig:hgnn_arch}
\end{figure}

Figure~\ref{fig:hgnn_arch} shows the architecture of the \hgnn. Note that \hgnn has exactly the same architecture as \lgnn and follows all the computations exactly in the same fashion until kinetic and potential energies. Once these energies are obtained, instead of computing the Lagrangian, the Hamiltonian of the system is computed using the equation $H = T + V$ where $T$ is total kinetic energy of system and $V$ is the total potential energy of the system. Finally, the acceleration is computed using the predicted Hamiltonian employing the appropriate Hamiltonian's equation of motion.

\section{Additional Experiments}
\label{app:add_exp}

\subsection{Data Efficiency}
\label{app:dataefficiency}
\begin{figure}[t]
\vspace{-0.20in}
    \centering
    \includegraphics[width=0.60\textwidth]{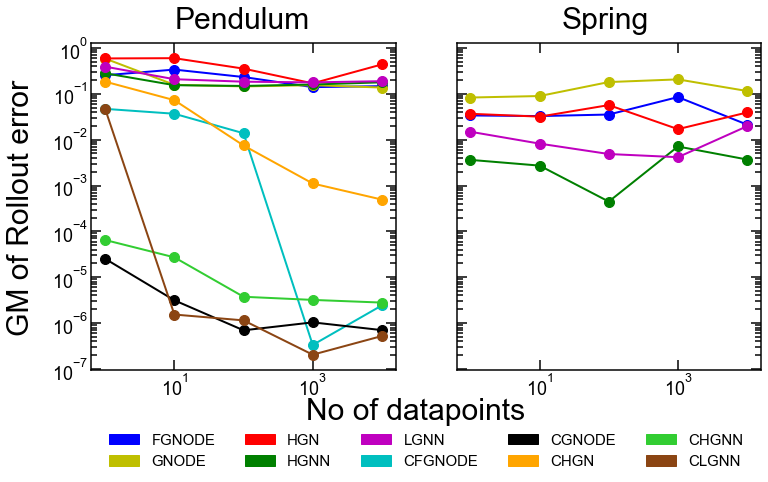}
    \vspace{-0.10in}
    \caption{Variation of rollout error against the number of data points used to train the models. }
    \label{fig:dataefficiency}
    \vspace{-0.20in}
\end{figure}

Figure~\ref{fig:dataefficiency}, shows the dataefficiency of the different models considers. Specifically, we evaluate the rollout error with respect to the number of data-points used to train each of the models. A clear trend emerges from this experiment. Specifically, in pendulum systems, we observe that models with explicit constraints significantly outperform their unconstrained counterparts with more data points by $\sim$5-6 orders of magnitude. In contrast, the unconstrained architectures show limited reduction in error in both spring and pendulum systems. This trend indicates that injecting explicit constraints in the model leads to more effective training. Further, we observe that the performance of \clgnn, \chgnn, and \cgnode are comparable for pendulum systems, while that of \chgn is poorer, despite having explicit constraints.

\begin{figure}[t]
    \centering
    \begin{subfigure}{0.49\textwidth}
        \centering
        \includegraphics[width=0.99\textwidth]{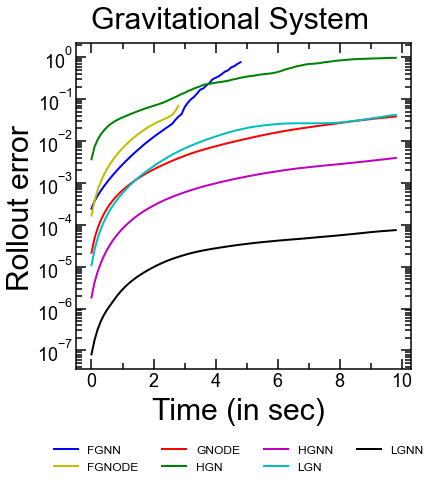} 
    \end{subfigure}\hfill
    \begin{subfigure}{0.49\textwidth}
        \centering
        \includegraphics[width=0.99\textwidth]{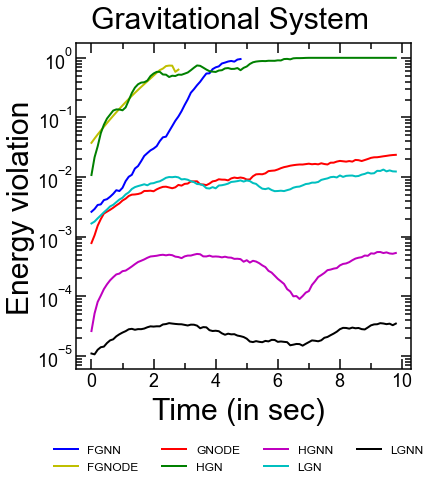}
    \end{subfigure}
    \caption{Rollout error and energy error for 4-body gravitational system with constraints with respect to time for \lgnn, \lgn \hgn, \hgnn, \gnode, \fgnode and \fgnn. The curve represents the average over 100 trajectories generated from random initial conditions.}
    \label{fig:grav_traj}
\end{figure}

\begin{figure}[t]
    \centering
    \begin{subfigure}{0.49\textwidth}
        \centering
        \includegraphics[width=0.99\textwidth]{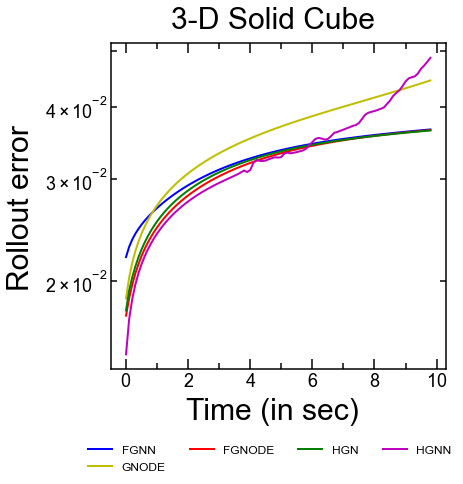} 
    \end{subfigure}\hfill
    \begin{subfigure}{0.49\textwidth}
        \centering
        \includegraphics[width=0.99\textwidth]{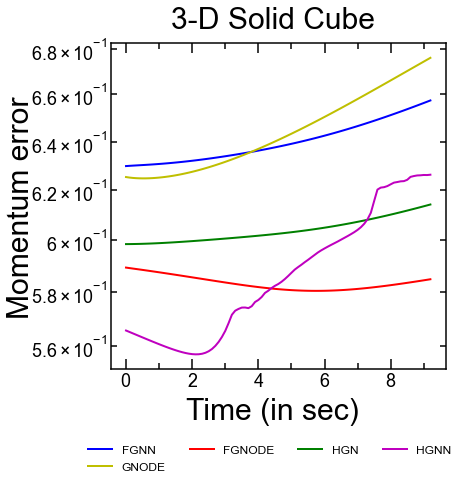}
    \end{subfigure}
    \caption{Rollout error and energy error for rigid body system with respect to time for \hgn, \hgnn, \gnode, \fgnode and \fgnn. The curve represents the average over 10 trajectories over the test dataset.}
    \label{fig:rigid_traj}
\end{figure}

\begin{figure}[t]
    \centering
    \begin{subfigure}{0.49\textwidth}
        \centering
        \includegraphics[width=0.99\textwidth]{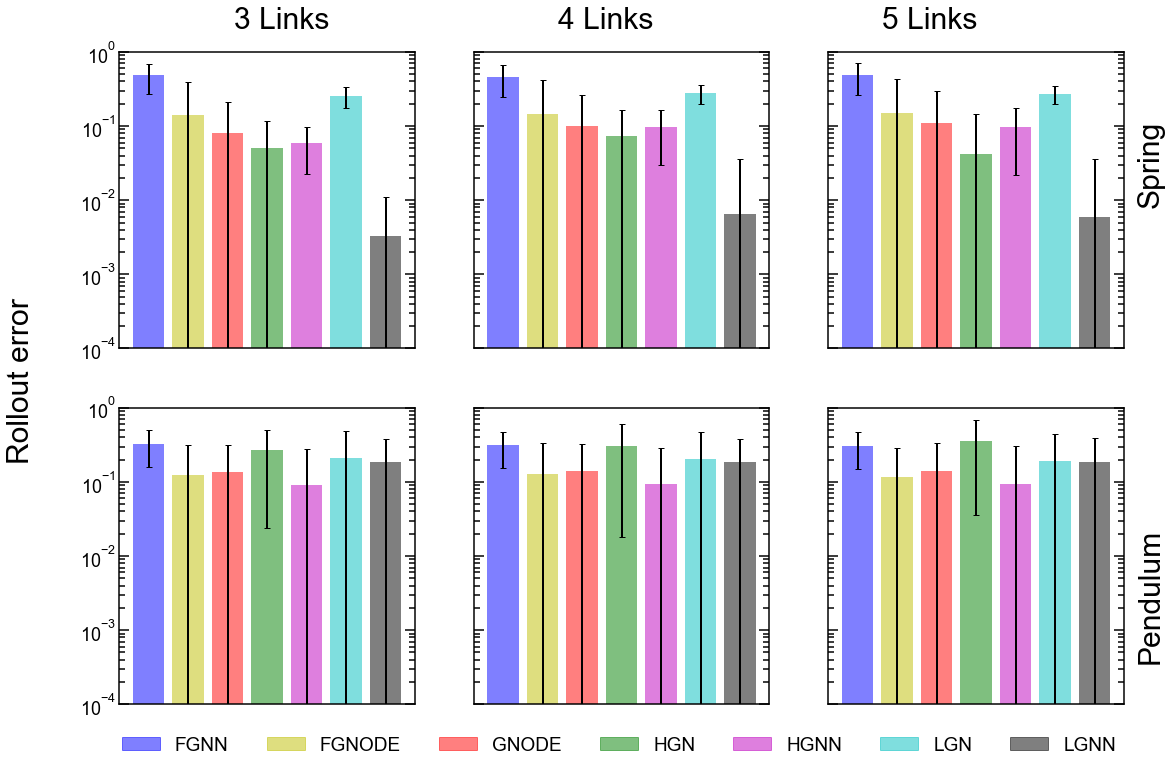} 
    \end{subfigure}\hfill
    \begin{subfigure}{0.49\textwidth}
        \centering
        \includegraphics[width=0.99\textwidth]{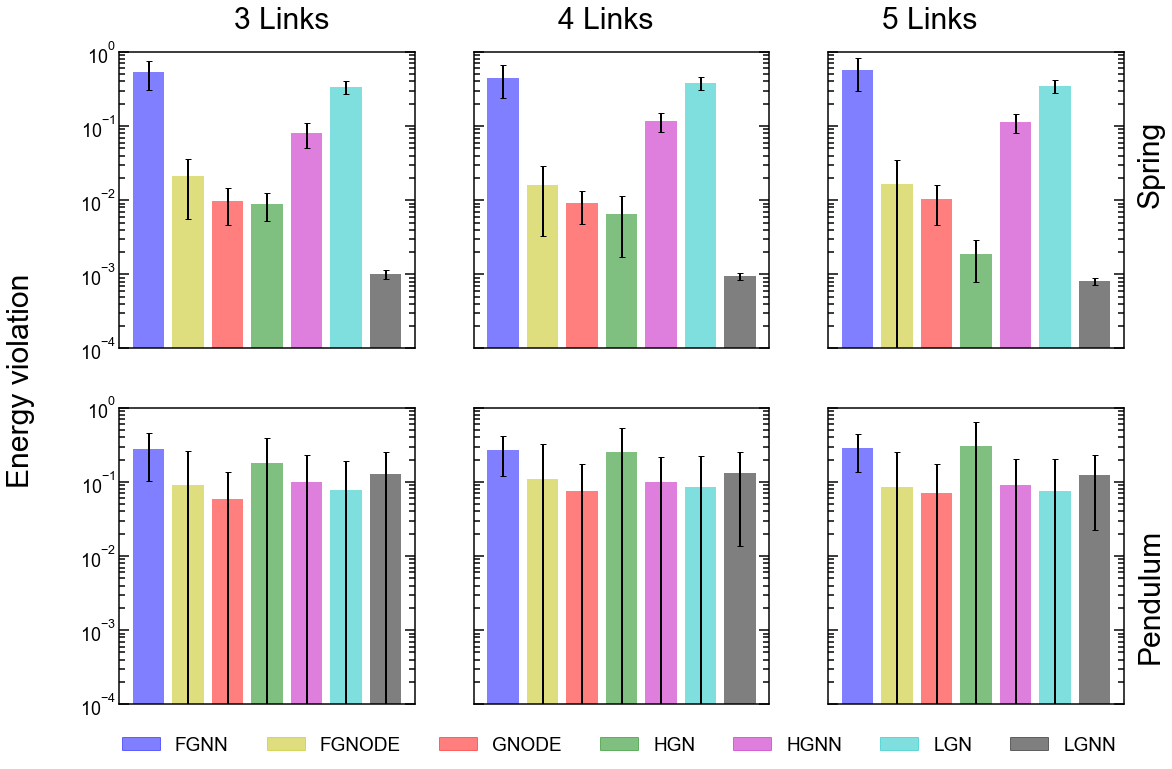} 
    \end{subfigure}
    \caption{Geometric mean of rollout error and energy error for 3-, 4-, 5-spring and 3-,4-,5-pendulum systems without constraints for \lgnn, \lgn \hgn, \hgnn, \gnode, \fgnode and \fgnn on noisy data. The error bar represents the 95\% confidence interval over 100 trajectories generated from random initial conditions.}
    \label{fig:noisy_bar}
\end{figure}

\begin{figure}[t]
    \centering
    \begin{subfigure}{0.49\textwidth}
        \centering
        \includegraphics[width=0.99\textwidth]{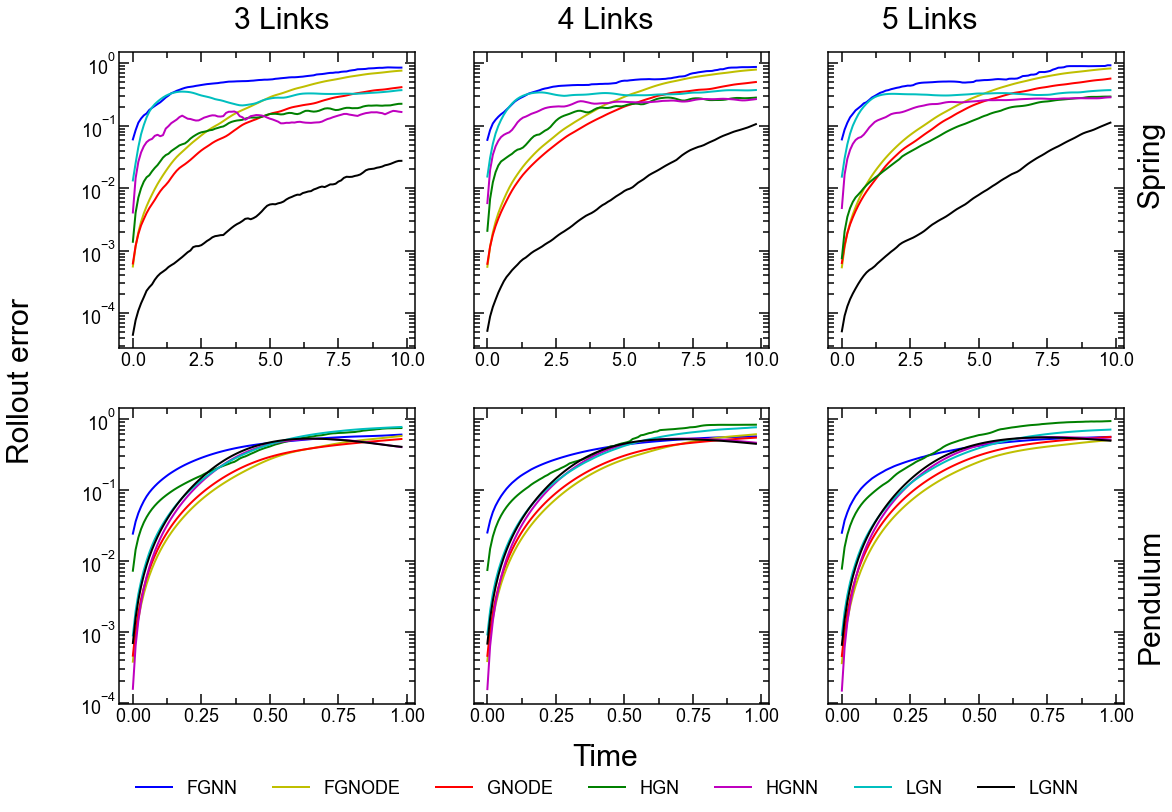} 
    \end{subfigure}\hfill
    \begin{subfigure}{0.49\textwidth}
        \centering
        \includegraphics[width=0.99\textwidth]{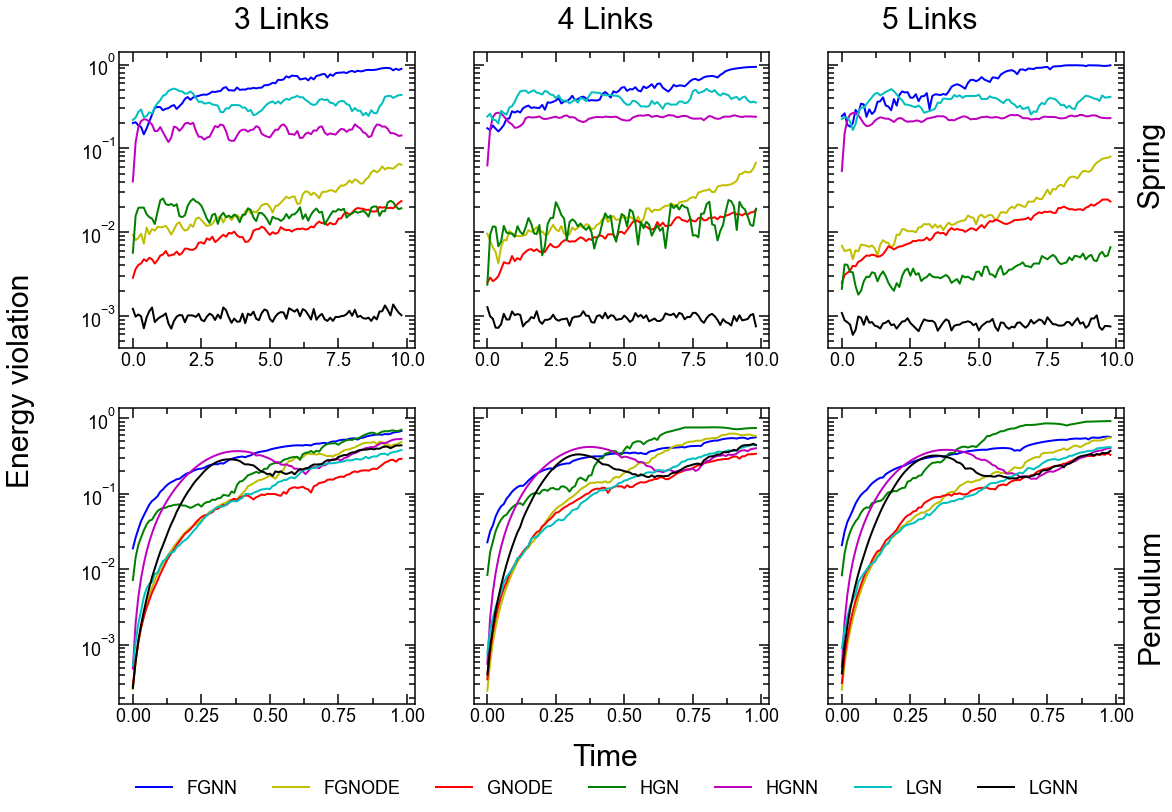} 
    \end{subfigure}
    \caption{ Rollout error and energy error for 3-, 4-, 5-spring and 3-, 4-, 5-pendulum systems without constraints with respect to time for \lgnn, \lgn \hgn, \hgnn, \gnode, \fgnode and \fgnn on noisy data. The curve represents the average over 100 trajectories generated from random initial conditions.}
    \label{fig:noisy_traj}
\end{figure}

\begin{figure}[t]
    \centering
    \begin{subfigure}{0.49\textwidth}
        \centering
        \includegraphics[width=0.99\textwidth]{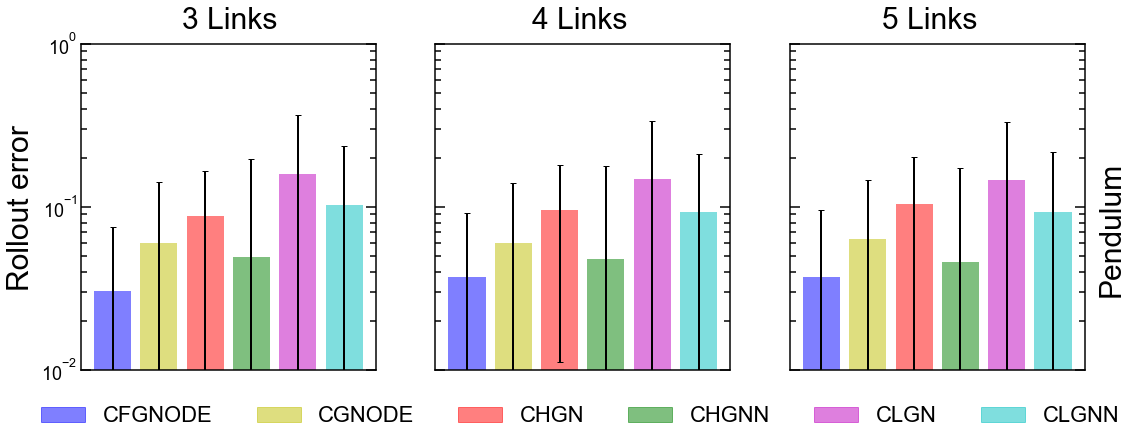} 
    \end{subfigure}\hfill
    \begin{subfigure}{0.49\textwidth}
        \centering
        \includegraphics[width=0.99\textwidth]{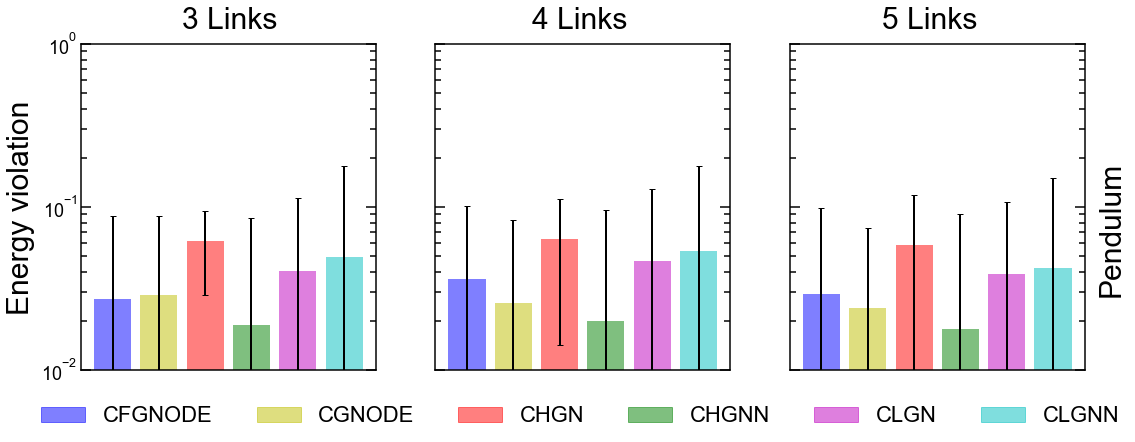} 
    \end{subfigure}
    \caption{Geometric mean of rollout error and energy error for 3-,4-,5-pendulum systems with constraints for \clgnn, \clgn \chgn, \chgnn, \cgnode, and \cfgnode on noisy data. The error bar represents the 95\% confidence interval over 100 trajectories generated from random initial conditions.}
    \label{fig:noisy_bar_constr}
\end{figure}

\begin{figure}[t]
\vspace{-0.10in}
    \centering
    \begin{subfigure}{0.49\textwidth}
        \centering
        \includegraphics[width=0.99\textwidth]{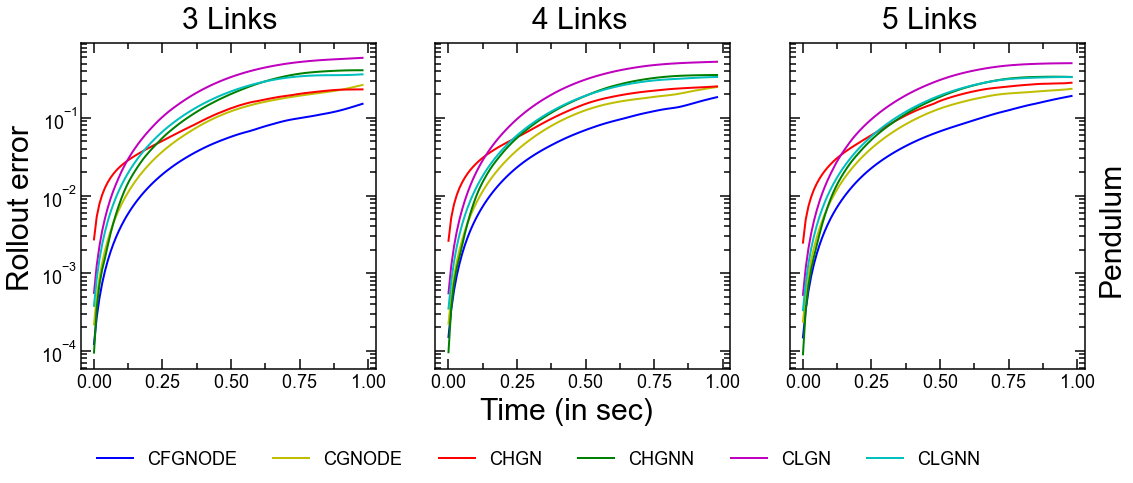} 
    \end{subfigure}\hfill
    \begin{subfigure}{0.49\textwidth}
        \centering
        \includegraphics[width=0.99\textwidth]{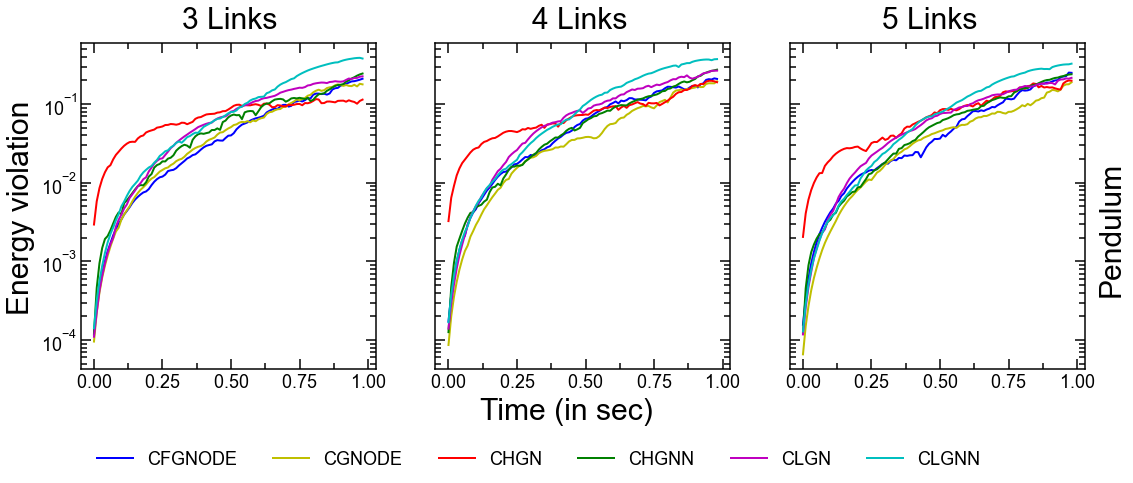}
    \end{subfigure}
    \caption{  Rollout error and energy error for 3-,4-,5-pendulum systems with constraints with respect to time for \clgnn, \clgn \chgn, \chgnn, \cgnode, and \cfgnode on noisy data. The curve represents the average over 100 trajectories generated from random initial conditions.}
    \label{fig:noisy_traj_constr}
\end{figure}


\subsection{Robustness to Noise}\label{app:noise}
\begin{figure}[t]
    \centering
    \includegraphics[width=0.9\textwidth]{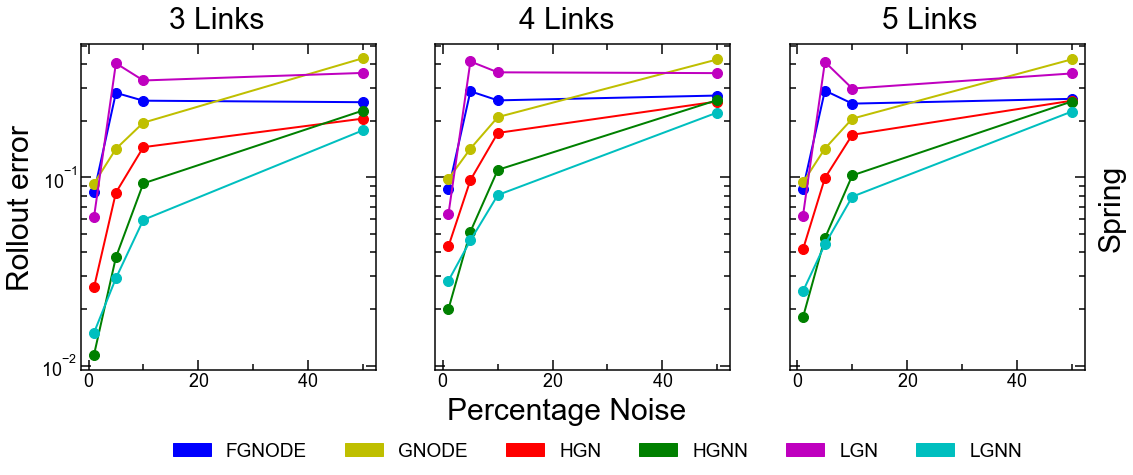}
    \caption{Rollout error with respect to percentage noise in the dataset used to train the 5-spring system.}
    \label{fig:noise_efficiency_rollout}
\end{figure}

\begin{figure}[t]
\vspace{-0.20in}
    \centering
    \includegraphics[width=0.90\textwidth]{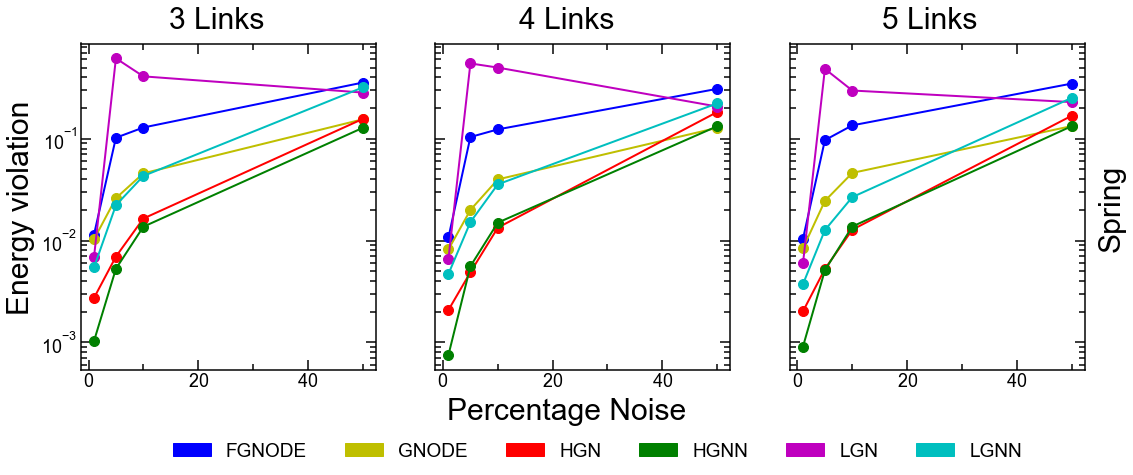}
    \caption{Energy error with respect to percentage of noise in the dataset used to train the 5-spring model.}
    \label{fig:noise_efficiency_energy}
\end{figure}

\vspace{-0.10in}
\rev{To evaluate robustness of the evaluated \gnns to  noise, we inject Gaussian noise to every data point in the training dataset with mean $0$ and standard deviation $1$. The forward simulation error is  calculated by comparing it to the ground truth trajectories, without adding any noise to those data points. Due to space limitations, the plots are in the appendix. Figure~\ref{fig:noisy_bar} shows the performance on unconstrained architectures, while Figure~\ref{fig:noisy_traj} shows the time evolution of energy and rollout error.
Figure~\ref{fig:noisy_bar_constr} and Figure~\ref{fig:noisy_traj_constr} analyze the same, respectively, on pendulum for constrained systems. Finally, we also show the variation of energy and rollout error for spring system for varying percentages of noise, namely, 1\%, 5\%, 10\% and 50\% of the standard deviation of the data (see Figs.~\ref{fig:noise_efficiency_rollout},\ref{fig:noise_efficiency_energy}.)} 

\rev{When compared to training on clean data, we observe that the trends remain similar. Specifically, \clgn and \chgn continue to be the poorest performers in the constrained setting. In the unconstrained setting, the same trend continues; \fgnn and \lgn continue exhibiting highest errors, while \hgn remains the best architecture. However, across architectures we observe almost a 10-fold increase in error. All in all, this experiment reveals that while the choice of architectures remain unaffected, all display reduced accuracy. Hence, enabling better robustness would be an important research direction to pursue.}
\vspace{-0.10in}

\section{Details of Experimental Setup}
\label{app:setup}

\vspace{+0.20in}
\subsection{Dataset generation}

\textbf{Software packages:} numpy-1.20.3, jax-0.2.24, jax-md-0.1.20, jaxlib-0.1.73, jraph-0.0.1.dev \\
\textbf{Hardware:}
Chip: Intel Xeon,
Total Number of Cores: 64, 
Memory: 128 GB, 
System OS: Ubuntu 18.04.5 LTS.

For all variants of \lgnn, \lgn, \gnode, \fgnode, \fgn: All the datasets are generated using the known Lagrangian of the pendulum and spring systems, along with the constraints, as described in Section~\ref{sec:experiments}. For each system, we create the training data by performing forward simulations with $100$ random initial conditions. For the pendulum system, a timestep of $10^{-5}s$ is used to integrate the equations of motion, while for the spring system, a timestep of $10^{-3}s$ is used. The velocity-Verlet algorithm is used to integrate the equations of motion due to its ability to conserve the energy in long trajectory integration. 

While for \hgn and \hgnn, datasets are generated using Hamiltonian mechanics. Runge-Kutta integrator is used to integrate the equations of motion due to the first order nature of the Hamiltonian equations in contrast to the second order nature of \lgnn and \gnode.

From the 100 simulations for pendulum and spring system, obtained from the rollout starting from 100 random initial conditions, 100 data points are extracted per simulation, resulting in a total of 10000 data points. Data points were collected every 1000 and 100 timesteps for the pendulum and spring systems, respectively. Thus, each training trajectory of the spring and pendulum systems are $10s$ and $1s$ long, respectively. Here, we do not train from the trajectory. Rather, we randomly sample different states from the training set to predict the acceleration.

\rev{For the gravitational system, the dataset is generated using the known Lagrangian of the gravitational system. We create the training data by performing forward simulation from a known stable state. We use a timestep of $10^{-3}s$, which is used to integrate the equations of motion, and similar to the pendulum and spring system, we use the velocity-verlet algorithm for the integration of the equations.}

\rev{The known stable state is simulated forward and datapoints were collected every 100 timesteps, for a total of 10000 datapoints. Similar to the pendulum and spring systems, we randomly sample different states from the training set to predict the acceleration.}

\rev{For 3D solid cube, we have generated ground truth data using peridynamics simulation, single initially compressed cube was relaxed for $100s$ with time step $0.1s$. Using this we have generated total 1000 data points.}

\subsection{Training details}
The training dataset is divided in 75:25 ratio randomly, where the 75\% is used for training and 25\% is used as the validation set. Further, the trained models are tested on its ability to predict the correct trajectory, a task it was not trained on. Specifically, the pendulum systems are tested for $1s$, that is $10^5$ timesteps, and spring systems for $20s$, that is $2\times10^4$ timesteps on 100 different trajectories created from random initial conditions. All models are trained for 10000 epochs. A learning rate of $10^{-3}$ was used with the Adam optimizer for the training.\\

\subsection{Loss function} 
\label{sec:loss_fn}
Based on the predicted $\xddot$, the positions and velocities are predicted using the \textit{Velocity Verlet} integration. The loss function is computed by using the predicted and actual accelerations at timesteps $2, 3,\ldots,\mathcal{T}$ in a trajectory $\mathbb{T}$, which is then back-propagated to train the MLPs. Specifically, the loss function is as follows.
\begin{equation}
    \label{eq:lossfunction}
  \mathcal{L}= \frac{1}{n}\left(\sum_{i=1}^n \left(\ddot{x}_i^{\mathbb{T},t}-\left(\hat{\ddot{x}}_i^{\mathbb{T},t}\right)\right)^2\right)
\end{equation}
 Here, $(\hat{\ddot{x}}_i^{\mathbb{T},t})$ is the predicted acceleration for the $i^{th}$ particle in trajectory $\mathbb{T}$ at time $t$ and $\ddot{x}_i^{\mathbb{T},t}$ is the true acceleration. $\mathbb{T}$ denotes a trajectory from $\mathfrak{T}$, the set of training trajectories. Note that the accelerations are computed directly from the ground truth trajectory using the Verlet algorithm as:
 \begin{equation}
\xddot(t)=\frac{1}{(\Delta t)^2}[\x(t+\Delta t)+\x(t-\Delta t)-2\x(t)]
\end{equation}
Since the integration of the equations of motion for the predicted trajectory is also performed using the same algorithm as: $\x(t+\Delta t)=2\x(t)-\x(t-\Delta t)+\xddot(\Delta t)^2$, this method is equivalent to training from trajectory/positions. 

\subsection{Hyper-parameters}
\label{app:hyper}
The default hyper-parameters used for training each architecture is provided below. 

$\bullet$\textbf{\gnode, \cgnode}
\begin{center}
\begin{tabular}{ |c|c| } 
 \hline
 \textbf{Parameter} & \textbf{Value} \\ 
 \hline
 Node embedding dimension & 5 \\ 
 Edge embedding dimension & 5 \\ 
 Hidden layer neurons (MLP) & 5 \\ 
 Number of hidden layers (MLP) & 2 \\ 
 Activation function & squareplus\\
 Number of layers of message passing & 1\\
 Optimizer & ADAM \\
 Learning rate & $1.0e^{-3}$ \\
 Batch size & 100 \\
 \hline
\end{tabular}
\end{center}

$\bullet$\textbf{\lgn, \clgn, \hgn, \fgnode, \cfgnode, \fgnn}
\begin{center}
\begin{tabular}{ |c|c| } 
 \hline
 \textbf{Parameter} & \textbf{Value} \\
 \hline
 Node embedding dimension & 8 \\ 
 Edge embedding dimension & 8 \\ 
 Hidden layer neurons (MLP) & 16 \\ 
 Number of hidden layers (MLP) & 2 \\ 
 Activation function & squareplus\\
 Number of layers of message passing & 1\\
 Optimizer & ADAM \\
 Learning rate & $1.0e^{-3}$ \\
 Batch size & 100 \\
 \hline
\end{tabular}
\end{center}

$\bullet$\textbf{\lgnn, \clgnn, \hgnn, \chgnn}
\begin{center}
\begin{tabular}{ |c|c| } 
 \hline
 \textbf{Parameter} & \textbf{Value} \\
 \hline
 Node embedding dimension & 5 \\ 
 Edge embedding dimension & 5 \\ 
 Hidden layer neurons (MLP) & 5 \\ 
 Number of hidden layers (MLP) & 2 \\ 
 Activation function & squareplus\\
 Number of layers of message passing(pendulum) & 2\\
  Number of layers of message passing(spring) & 1\\
 Optimizer & ADAM \\
 Learning rate & $1.0e^{-3}$ \\
 Batch size & 100 \\
 \hline
\end{tabular}
\end{center}

\color{black}

\subsection{Hyper-parameter Search}
\label{app:hypersearch}
\color{black}
$\bullet$ \textbf{Learning rate}

\fgnn \\
\begin{center}
\begin{tabular}{ |c|c|c| } 
\hline
\textbf{LR Value} & \textbf{Geometric mean of Zerr} & \textbf{Time (in sec)} \\
\hline
0.001 & 0.0852 & 4226\\
0.003 & 0.0616 & 4394\\
0.01 & 0.1219 &  4376\\
0.03 & 0.0962 & 4436\\ 
0.1 & 0.0846 & 4476\\
0.3 & 0.0628 & 4557\\
\hline
\end{tabular}
\end{center}

\gnode \\
\begin{center}
\begin{tabular}{ |c|c|c| } 
\hline
\textbf{LR Value} & \textbf{Geometric mean of Zerr} & \textbf{Time (in sec)} \\
\hline
0.001 & 0.1371 & 5189 \\
0.003 & 0.1507 & 5242 \\
0.01 & 0.1288 & 5250 \\
0.03 & 0.1267 & 5176 \\
0.1 & 0.1284 & 5206 \\
0.3 & 0.1491 & 5179 \\
\hline
\end{tabular}
\end{center}

\hgnn \\
\begin{center}
\begin{tabular}{ |c|c|c| } 
\hline
\textbf{LR Value} & \textbf{Geometric mean of Zerr} & \textbf{Time (in sec)} \\
\hline
0.001 & 0.1868 & 3856 \\
0.003 & 0.1895 & 3613 \\
0.01 & 0.1622 & 3878 \\
0.03 & 0.1878 & 4089 \\
0.1 & 0.146 & 9195 \\
0.3 & 0.2356 & 6859 \\
\hline
\end{tabular}
\end{center}

\lgnn
\begin{center}
\begin{tabular}{ |c|c|c| } 
\hline
\textbf{LR Value} & \textbf{Geometric mean of Zerr} & \textbf{Time (in sec)} \\
\hline
0.001 & 0.1939 & 38082 \\
0.003 & 0.189 & 11904 \\
0.01 & 0.1869 & 11890 \\
0.03 & 0.1879 & 12314 \\
0.1 & 0.2316 & 11845 \\
0.3 & 0.279 & 12320 \\
\hline
\end{tabular}
\end{center}

$\bullet$ \textbf{Number of message-passing layers}

\fgnn \\
\begin{center}
\begin{tabular}{ |c|c|c| } 
\hline
\textbf{No of message passing layers} & \textbf{Geometric mean of Zerr} & \textbf{Time (in sec)} \\
\hline
1 & 0.0852 & 6234 \\
2 & 1.0 & 7552 \\
3 & 0.1724 & 8024 \\
4 & 0.0786 & 8455 \\
\hline
\end{tabular}
\end{center}

\gnode \\
\begin{center}
\begin{tabular}{ |c|c|c| } 
\hline
\textbf{No of message passing layers} & \textbf{Geometric mean of Zerr} & \textbf{Time (in sec)} \\
\hline
1 & 0.1491 & 7833 \\
2 & 0.1649 & 6850 \\
3 & 0.1721 & 6924 \\
4 & 0.2078 & 7429 \\
\hline
\end{tabular}
\end{center}

\hgnn \\
\begin{center}
\begin{tabular}{ |c|c|c| } 
\hline
\textbf{No of message passing layers} & \textbf{Geometric mean of Zerr} & \textbf{Time (in sec)} \\
\hline
1 & 0.1782 & 8924 \\
2 & 0.1796 & 8254 \\
3 & 0.1583 & 9095 \\
4 & 0.1608 & 11199 \\
\hline
\end{tabular}
\end{center}

\lgnn
\begin{center}
\begin{tabular}{ |c|c|c| } 
\hline
\textbf{No of message passing layers} & \textbf{Geometric mean of Zerr} & \textbf{Time (in sec)} \\
\hline
1 & 0.1887 & 11649 \\
2 & 0.2319 & 12345 \\
3 & 0.1903 & 12172 \\
4 & 0.1839 & 12838 \\
\hline
\end{tabular}
\end{center}

$\bullet$ \textbf{Number of hidden layers in MLP}

\fgnn \\
\begin{center}
\begin{tabular}{ |c|c|c| } 
\hline
\textbf{No of hidden layers} & \textbf{Geometric mean of Zerr} & \textbf{Time (in sec)} \\
\hline
5 & 0.1132 & 5423 \\
10 & 0.0848 & 5926 \\
15 & 0.0821 & 6583 \\
25 & 0.0934 & 7446 \\
\hline
\end{tabular}
\end{center}

\gnode \\
\begin{center}
\begin{tabular}{ |c|c|c| } 
\hline
\textbf{No of hidden layers} & \textbf{Geometric mean of Zerr} & \textbf{Time (in sec)} \\
\hline
5 & 0.2078 & 25559 \\
10 & 0.1697 & 9246 \\
15 & 0.2084 & 9728 \\
25 & 0.185 & 10747 \\
\hline
\end{tabular}
\end{center}

\hgnn \\
\begin{center}
\begin{tabular}{ |c|c|c| } 
\hline
\textbf{No of hidden layers} & \textbf{Geometric mean of Zerr} & \textbf{Time (in sec)} \\
\hline
5 & 0.1782 & 8984 \\
10 & 0.1524 & 11307 \\
15 & 0.1543 & 12605 \\
25 & 0.1497 & 17498 \\
\hline
\end{tabular}
\end{center}

\lgnn
\begin{center}
\begin{tabular}{ |c|c|c| } 
\hline
\textbf{No of hidden layers} & \textbf{Geometric mean of Zerr} & \textbf{Time (in sec)} \\
\hline
5 & 0.1887 & 12885 \\
10 & 0.18 & 14954 \\
15 & 0.1692 & 16533 \\
25 & 0.1731 & 21011 \\
\hline
\end{tabular}
\end{center}

$\bullet$ \textbf{Embedding dimensionality in hidden layers of MLP}

\fgnn \\
\begin{center}
\begin{tabular}{ |c|c|c| } 
\hline
\textbf{No of Neurons} & \textbf{Geometric mean of Zerr} & \textbf{Time (in sec)} \\
\hline
1 & 0.0646 & 3973 \\
2 & 0.1132 & 5462 \\
4 & 0.0763 & 8258 \\
8 & 0.075 & 13900 \\
\hline
\end{tabular}
\end{center}

\gnode \\
\begin{center}
\begin{tabular}{ |c|c|c| } 
\hline
\textbf{No of Neurons} & \textbf{Geometric mean of Zerr} & \textbf{Time (in sec)} \\
\hline
1 & 0.187 & 7664 \\
2 & 0.2078 & 9895 \\
4 & 0.1869 & 12258 \\
8 & 0.7755 & 20041 \\
\hline
\end{tabular}
\end{center}

\hgnn \\
\begin{center}
\begin{tabular}{ |c|c|c| } 
\hline
\textbf{No of Neurons} & \textbf{Geometric mean of Zerr} & \textbf{Time (in sec)} \\
\hline
1 & 0.1886 & 7689 \\
2 & 0.1782 & 8984 \\
4 & 0.1555 & 12615 \\
8 & 0.1553 & 19389 \\
\hline
\end{tabular}
\end{center}

\lgnn
\begin{center}
\begin{tabular}{ |c|c|c| } 
\hline
\textbf{No of Neurons} & \textbf{Geometric mean of Zerr} & \textbf{Time (in sec)} \\
\hline
1 & 0.2351 & 10448 \\
2 & 0.1887 & 13070 \\
4 & 0.18 & 15710 \\
8 & 0.1674 & 24039 \\
\hline
\end{tabular}
\end{center}

$\bullet$ \textbf{Activation function}

\fgnn \\
\begin{center}
\begin{tabular}{ |c|c|c| } 
\hline
\textbf{Activation Function} & \textbf{Geometric mean of Zerr} & \textbf{Time (in sec)} \\
\hline
softplus & 0.886 & 6141 \\
squareplus & 0.1757 & 1340 \\
\hline
\end{tabular}
\end{center}

\gnode \\
\begin{center}
\begin{tabular}{ |c|c|c| } 
\hline
\textbf{Activation Function} & \textbf{Geometric mean of Zerr} & \textbf{Time (in sec)} \\
\hline
softplus & 0.1287 & 6970 \\
squareplus & 0.1371 & 6425 \\
\hline
\end{tabular}
\end{center}

\hgnn \\
\begin{center}
\begin{tabular}{ |c|c|c| } 
\hline
\textbf{Activation Function} & \textbf{Geometric mean of Zerr} & \textbf{Time (in sec)} \\
\hline
softplus & 0.1953 & 9685 \\
squareplus & 0.1782 & 8463 \\
\hline
\end{tabular}
\end{center}

\lgnn
\begin{center}
\begin{tabular}{ |c|c|c| } 
\hline
\textbf{Activation Function} & \textbf{Geometric mean of Zerr} & \textbf{Time (in sec)} \\
\hline
softplus & 0.1921 & 11620 \\
squareplus & 0.1887 & 38082 \\
\hline
\end{tabular}
\end{center}

\section{Training and Simulation Time}
\label{app:time}
The key insight obtained from Tables~\ref{tab:pendulumtime}-\ref{tab:rigidtime} is that the \lgn family of architectures take significantly longer to train. The \gnode family is marginally faster on average than the \hgn family. The \lgn family is the slowest since the lagrangian needs to be differentiated, which leads to the differentiation over the \gnn parameters. In \hgn, the \gnn outputs Hamiltonian and thus there is only one layer of differentiation to learn the \gnn parameters. Finally, in \gnode, the output is only integrated. Thus, to summarize,  in \lgn family, the order of differentiation is double, in \hgn the order is single and in \gnode, the order is zero.

\begin{table}[h]
\centering
\begin{tabular}{ |l|c|c| }
\toprule
\textbf{Models} & \textbf{Training time (in sec)} & \textbf{Forward Simulation time (in sec)} \\
\midrule
\cfgnode & 9475 & 2.21\\
\cgnode & 8784 & 1.57\\
\clgn & 55810 & 16.97\\
\clgnn & 27614 & 3.85\\
\chgn & 6130 & 0.54\\
\chgnn & 11038 & 0.82\\
\fgnn & 1325 & 0.02\\
\fgnode & 8097 & 1.75\\
\gnode & 6341 & 1.02\\
\lgn & 55042 & 14.82\\
\lgnn & 37752 & 13.53\\
\hgn & 2512 & 0.76\\
\hgnn & 8365 & 0.51\\
\bottomrule
\end{tabular}
\caption{\rev{Training and inference times in pendulum systems.}}
\label{tab:pendulumtime}
\end{table}

\begin{table}[h]
\centering
\begin{tabular}{ |l|c|c| }
\toprule
\textbf{Models} & \textbf{Training time (in sec)} & \textbf{Forward Simulation time (in sec)} \\
\midrule

\fgnn & 1738 & 0.01\\
\fgnode & 1337 & 0.25\\
\gnode & 6977 & 0.10\\
\lgn & 141962 & 5.81\\
\lgnn & 25710 & 0.81\\
\hgn & 14053 & 0.43\\
\hgnn & 18128 & 0.89\\
\bottomrule
\end{tabular}
\caption{\rev{Training and inference times in spring systems.}}
\label{tab:springtime}
\end{table}

\begin{table}[h]
\centering
\begin{tabular}{|l|c|c| }
\toprule
\textbf{Models} & \textbf{Training time (in sec)} & \textbf{Forward Simulation time (in sec)} \\
\midrule
\fgnn & 13136 & 0.30\\
\fgnode & 13238 & 12.94\\
\gnode & 11422 & 8.58\\
\hgn & 56996 & 5.21\\
\hgnn & 46038 & 8.04\\
\bottomrule
\end{tabular}
\caption{\rev{Training and inference times in rigid body systems.}}
\label{tab:rigidtime}
\end{table}
\color{black}
\end{document}